\documentclass[sigconf, acmsmall, screen, authorversion]{acmart}

\AtBeginDocument{%
  }

\setcopyright{acmlicensed}
\acmJournal{PACMCGIT}
\acmYear{2025} \acmVolume{8} \acmNumber{1} \acmArticle{19} \acmMonth{5}\acmDOI{10.1145/3728305}
\usepackage{acmart-taps}
\def\toprule{\hline}
\def\midrule{\hline}
\def\bottomrule{\hline}

\begin{document}

\title{ESCT3D: Efficient and Selectively Controllable Text-Driven 3D Content Generation with Gaussian Splatting}


\author{Huiqi Wu}
\email{wuhuiqi@seu.edu.cn}
\affiliation{%
  \institution{Southeast University}
  \city{Nanjing}
  \country{China}}

\author{Jianbo Mei}
\email{230248990@seu.edu.cn}
\affiliation{%
  \institution{Southeast University}
  \city{Nanjing}
  \country{China}}

\author{Yingjie Huang}
\email{220236705@seu.edu.cn}
\affiliation{%
  \institution{Southeast University}
  \city{Nanjing}
  \country{China}}

\author{Yining Xu}
\email{yiningxu@seu.edu.cn}
\affiliation{%
  \institution{Southeast University}
  \city{Nanjing}
  \country{China}}

\author{Jingjiao You}
\email{220224308@seu.edu.cn}
\affiliation{%
  \institution{Southeast University}
  \city{Nanjing}
  \country{China}}

\author{Yilong Liu}
\email{220234735@seu.edu.cn}
\affiliation{%
  \institution{Southeast University}
  \city{Nanjing}
  \country{China}}

\author{Li Yao}
\authornote{Corresponding author}
\email{yao.li@seu.edu.cn}
\affiliation{%
  \institution{Southeast University}
  \city{Nanjing}
  \state{Jiangsu}
  \country{China}
}

\renewcommand{\shortauthors}{Wu et al.}

\begin{abstract}
In recent years, significant advancements have been made in text-driven 3D content generation. However, several challenges remain. In practical applications, users often provide extremely simple text inputs while expecting high-quality 3D content. Generating optimal results from such minimal text is a difficult task due to the strong dependency of text-to-3D models on the quality of input prompts. Moreover, the generation process exhibits high variability, making it difficult to control. Consequently, multiple iterations are typically required to produce content that meets user expectations, reducing generation efficiency. To address this issue, we propose GPT-4V for self-optimization, which significantly enhances the efficiency of generating satisfactory content in a single attempt. Furthermore, the controllability of text-to-3D generation methods has not been fully explored. Our approach enables users to not only provide textual descriptions but also specify additional conditions, such as style, edges, scribbles, poses, or combinations of multiple conditions, allowing for more precise control over the generated 3D content. Additionally, during training, we effectively integrate multi-view information, including multi-view depth, masks, features, and images, to address the common Janus problem in 3D content generation. Extensive experiments demonstrate that our method achieves robust generalization, facilitating the efficient and controllable generation of high-quality 3D content.
\end{abstract}


\begin{teaserfigure}
  \includegraphics[width=\textwidth]{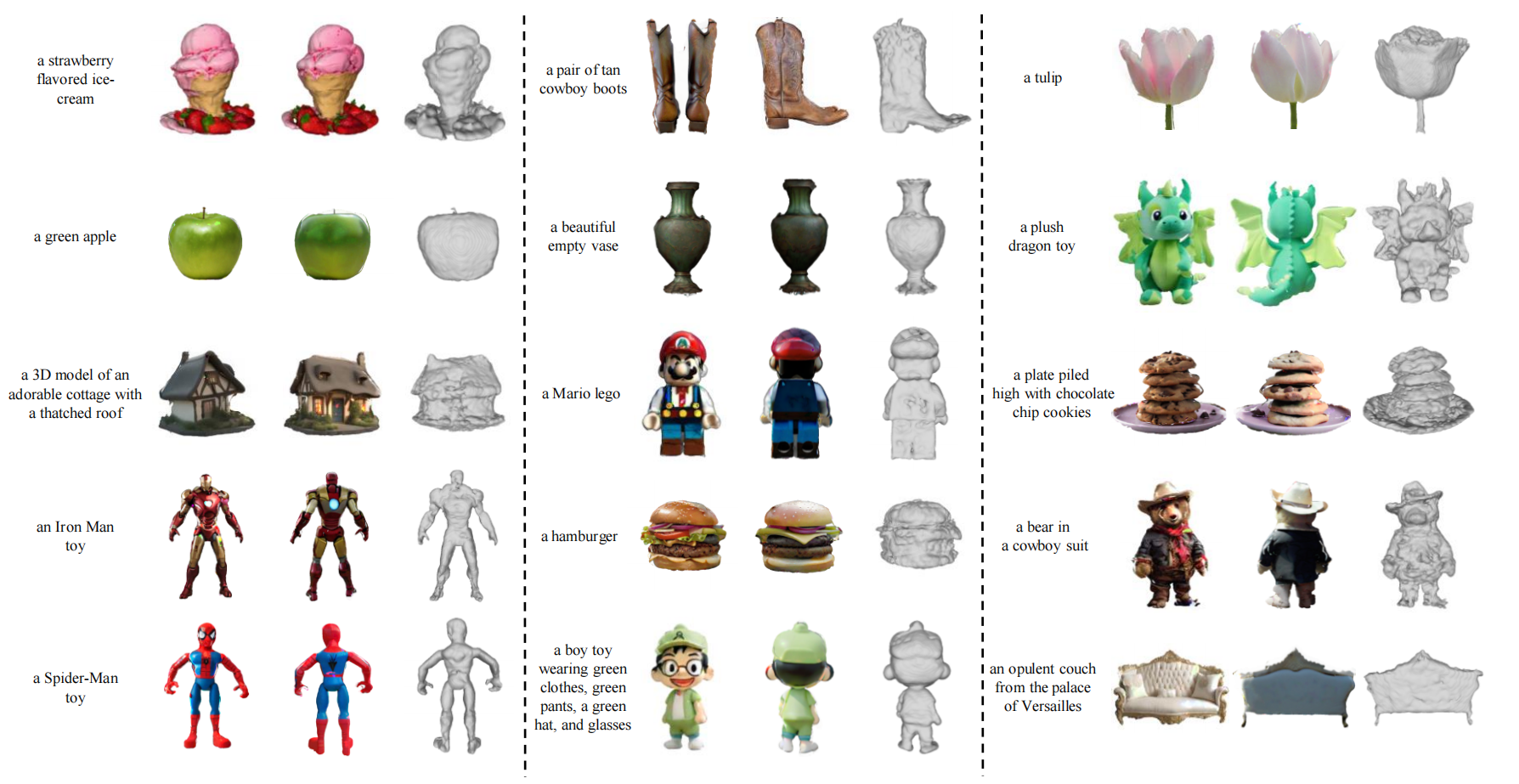}
  \caption{ESCT3D presents an efficient and selectively controllable approach for generating high-quality 3D content with minimal user input.}
  \label{fig:teaser}
\end{teaserfigure}

\maketitle

\section{Introduction}
The continuous advancement of artificial intelligence is fundamentally transforming the generation of 3D content. In recent years, 3D content generation has emerged as a significant research focus, offering substantial benefits to various industries, such as gaming, virtual reality, and film production. However, traditional 3D content creation processes are time-consuming and expensive. Therefore, the development of efficient 3D content generation technologies is essential for reducing entry barriers, optimizing workflows, and enhancing the overall efficiency of 3D content creation, thereby enabling broader participation in the creation of 3D content.

Recent advancements in the field of 2D image generation have significantly accelerated research in 3D generation tasks. These advancements can be largely attributed to the favorable characteristics of image diffusion models \cite{liu2023zero, rombach2022high} and differentiable 3D representations \cite{kerbl20233d, mildenhall2021nerf, shen2021deep, wang2021neus}. In particular, recent methods based on Score Distillation Sampling (SDS) \cite{poole2022dreamfusion} attempt to extract 3D knowledge from pre-trained large text-to-image generation models \cite{liu2023zero, rombach2022high, shi2023mvdream}, achieving impressive results \cite{chen2023fantasia3d, lin2023magic3d, metzer2023latent, poole2022dreamfusion, tang2023dreamgaussian, tang2023make, wang2024prolificdreamer}.

Although many works focus on text-to-3D tasks, an important but yet insufficiently explored area lies in efficient and selectively controllable text-to-3D generation. In this work, we propose a novel 3D generation pipeline where users can not only provide simple text but also additional conditions, such as style, edges, scribbles, poses, or combinations of multiple conditions, to control the generated content. At the same time, based on previous works \cite{hao2023optimizing, yang2023idea2img, wang2024promptcharm}, our framework introduces a novel self-optimization process that enables the input of simple text, which is then refined to generate more expressive prompts better suited for 3D content generation. This significantly improves both the quality and efficiency of the generation process. 
Compared to previous works \cite{tang2023dreamgaussian, yi2024gaussiandreamer}, our framework, for the first time, integrates multi-view depth, masks, features, and images simultaneously. This integration significantly improves 3D generation results, effectively addressing the challenges of insufficient detail in novel viewpoints and the Janus problem in text-based 3D content generation.
Overall, our approach can efficiently generate high-fidelity content while reducing the reliance on extensive manual work, paving the way for more user-friendly 3D content generation. 

Our contributions can be summarized as follows.
\begin{itemize}
\item We propose a novel controllable text-to-3D content generation method with multi-modal self-optimization, which can refine the input text into richer prompts and optionally integrate additional conditions beyond the text to enable the generation of controllable 3D results.
\item We combine multi-view information, including multi-view depth, masks, features, and images, as constraints to ensure multi-view consistency and enrich the details of other views.
\item We achieve the efficient generation of realistic 3D content. Our approach effectively balances optimization time and generation fidelity, offering new potential for the practical application of 3D content generation.
\end{itemize}

\section{Related Work}

\subsection{3D Representation Method}
Representative representation methods for 3D contents include \cite{gao2022get3d, chen2023fantasia3d, li2023sweetdreamer, lin2023magic3d}, which use explicit optimizable mesh representation methods, as well as other methods such as using point clouds \cite{nichol2022point, vahdat2022lion} and meshes \cite{liu2023meshdiffusion} as generation methods for 3D representations. These are widely used because they allow for direct and intuitive control of each element, and enable fast rendering through rasterization pipelines. However, due to their simple structure, they require a large number of elements for detailed expression. In order to express more effectively, complex primitives have been developed, such as rectangular prisms \cite{tulsiani2017learning}, gaussian prisms \cite{foygel2010extended}, ellipsoids \cite{genova2019learning}, hyperquadric surfaces \cite{paschalidou2019superquadrics}, convex hull \cite{deng2020cvxnet}, and polynomial surfaces \cite{yavartanoo20213dias}. Although graphic elements enhance the expressive power of complex geometry, they are still difficult to represent real 3D scenes due to their simple color representation. The differentiable representation of 3D scenes is crucial for representing various scenes and achieving subsequent tasks such as text to 3D generation. There are already some works that use neural networks as implicit representations to express more detailed 3D scenes. They train neural networks to express scenes that create desired attributes in 3D coordinates, such as signed distance functions \cite{park2019deepsdf}, RGB\( \alpha \) \cite{sitzmann2020implicit}.

In recent years, neural radiation fields (NeRF) \cite{mildenhall2021nerf} have made remarkable progress in the fields of computer vision and graphics. It is a deep learning method used to reconstruct 3D scenes and synthesize new viewpoints. It uses large MLPs to represent 3D scenes and renders them through volume rendering. However, scenarios stored implicitly in network form are difficult to handle, and their training and inference speeds are slow. Subsequent work based on NeRF aims to improve reconstruction quality \cite{barron2021mip}, training speed \cite{muller2022instant}, represent large-scale scenes \cite{tancik2022block}, and achieve reconstruction under constrained conditions \cite{ahn2023neural}. Many methods in text to 3D content generation also use NeRF or its variants \cite{barron2021mip} as representation methods. However, when the input information is limited, NeRF has a lower processing speed and consumes more memory, resulting in higher time and computational costs, which limits its practical application. The demand for faster and more efficient rendering techniques has become increasingly evident, especially in applications requiring real-time performance, where it is clear that NeRF can no longer meet these needs. 3D Gaussian Splatting \cite{kerbl20233d} addresses this demand by introducing a novel representation technique that utilizes millions of 3D Gaussians. It has achieved great success in fast, high-quality reconstruction and has demonstrated advantages in preserving high-frequency details.

\subsection{3D Content Generation}
Drawing inspiration from the latest breakthroughs in 2D content generation \cite{chen2024text, tang2023dreamgaussian, rombach2022high}, the field of 3D content creation has experienced rapid development. The latest research on 3D content generation can be divided into two main categories: 3D native methods based on 3D datasets and lifting 2D diffusion model to 3D.

3D native methods \cite{nichol2022point, jun2023shap, gupta20233dgen} demonstrate the potential to generate 3D consistent assets within seconds, at the cost of extensive training on large-scale 3D datasets. Creating such a dataset requires a significant amount of time and labor costs. On the other hand, Dreamfusion \cite{poole2022dreamfusion} proposed Score Distillation Sampling (SDS) to address the limitations of 3D datasets by extracting 3D geometry and appearance from powerful 2D diffusion models \cite{saharia2022photorealistic}, which inspired the development of 2D enhancement methods \cite{chen2023fantasia3d, lin2023magic3d, wang2024prolificdreamer}.

{\bfseries 3D Pretrained Diffusion Model.} The main difference between 3D pretrained diffusion model and 2D diffusion model lifting to 3D lies in whether the training data used is 2D or 3D. The 3D diffusion model \cite{nichol2022point, jun2023shap, gupta20233dgen, gao2022get3d} is a pretrained model on text-3D paired data. After pretraining, they only need to reason to generate 3D assets, and models such as Point-E \cite{nichol2022point} and Shape-E \cite{jun2023shap} can generate 3D assets in minutes. But the reason why text to image generation is so successful is because they are driven by diffusion models trained on billions of images, and applying this method to 3D content generation will require a large-scale label dataset, which is difficult.

{\bfseries Lifting 2D Diffusion Model to 3D.} Inspired by the early success of Generative Adversarial Networks (GANs) in image generation \cite{goodfellow2014generative}, similar attempts have been made in 3D creation. However, the instability of GAN learning and memory limitations in 3D representation have limited the quality of generation. Encouraged by the recent success of diffusion models \cite{ho2020denoising} in the field of image generation, these models have adapted to complex cues and generated high-quality images and scenes closely related to the given cues \cite{avrahami2023blended, ho2022classifier, rombach2022high, ruiz2023dreambooth}. With the advancement of image generation, people's interest in 3D generation has surged, and there have been many attempts to introduce diffusion models into 3D representation.

In the text to 3D asset generation method \cite{chen2024it3d, lorraine2023att3d, park2023ed, raj2023dreambooth3d, seo2023let, song2023roomdreamer, wu2024hd}, besides using 3D pretrained diffusion models, upgrading 2D diffusion models to 3D is a training free approach. In addition, due to the abundance of two-dimensional image data, the assets generated by this method have higher diversity and fidelity. The model \cite{jain2022zero, michel2022text2mesh, mohammad2022clip, wang2022clip, sanghi2022clip} uses CLIP to align each view of the 3D representation model with the text, optimizing the 3D model by maximizing the text image alignment score. DreamField \cite{jain2022zero} trains 3D representations under CLIP guidance to achieve text to 3D content generation. However, due to the weak supervision of CLIP loss, the results were not ideal. 

With the advancement of diffusion models, DreamFusion \cite{poole2022dreamfusion} first proposed the SDS method, which allows training 3D models based on 2D knowledge of diffusion models, that is, using 2D diffusion models to update 3D representation models. It uses SDS loss to optimize the 3D model, matching the rendering of its random viewpoints with the text conditional image distribution of the powerful Text to Image diffusion model. 2D generative models are typically trained on large-scale image-text datasets that encompass a wide range of styles, objects, and scenes, significantly enhancing their generalization ability and enabling them to generate diverse content. DreamFusion inherits the creativity and generalization ability of 2D generative models and is capable of generating highly creative 3D assets. In order to address issues such as oversaturation and blurring, some works have adopted phased optimization strategies or proposed improved fractional distillation losses, thereby enhancing the realism of the photos. Subsequent methods \cite{chen2023fantasia3d, li2023sweetdreamer, lin2023magic3d, sun2023dreamcraft3d} were all based on DreamFusion, further improving the quality of 3D generation. Among them, ProlificDreamer \cite{wang2024prolificdreamer} introduced variational distillation to improve the low diversity of SDS generation content; Fantasia3D \cite{chen2023fantasia3d} uses physics based rendering to decouple geometry and materials to create high-quality meshes; SweetDreamer \cite{li2023sweetdreamer} improves consistency by aligning the 2D geometric priors in the diffusion model with clearly defined 3D shapes during the lifting process; Magic3D \cite{lin2023magic3d} improved the qualitative results by using two-stage training; DreamCraft3D \cite{sun2023dreamcraft3d} guides the geometric constraint and texture enhancement stages by utilizing 2D reference images; DreamGaussian \cite{tang2023dreamgaussian} based on 3D Gaussian Splatting provides an effective two-stage framework for text driven and image driven 3D generation. Compared to previous methods, we propose an efficient text-to-3D content generation approach that can self-optimize the input text into more informative prompts and optionally integrate additional conditions beyond textual input to enable the generation of controllable 3D results. Furthermore, we incorporate multi-view information, including multi-view depth, masks, features, and images, as constraints to ensure multi-view consistency and enhance the details of other views.
\begin{figure*}[htbp]
  \centering
  \includegraphics[width=\textwidth]{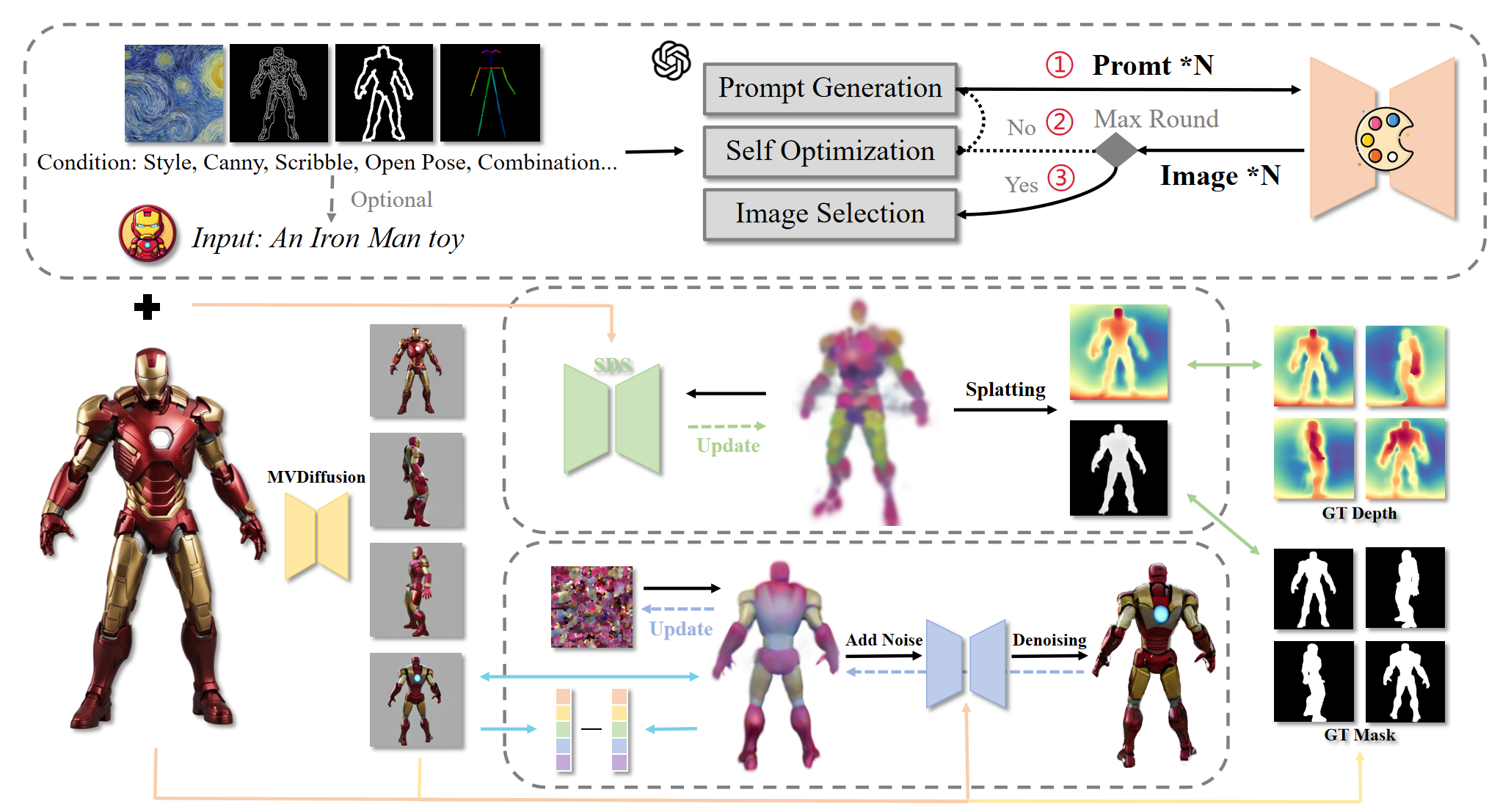}
  \caption{Framework. The multi-modal self-optimization framework is designed to control the generation of high-quality images. Then, 3D Gaussians are employed to effectively initialize geometry and appearance using SDS loss, while incorporating multi-view depth and mask information. Afterward, we iteratively refine the texture image using various multi view losses.}
  \Description{Framework}
  \label{fig:framework}
\end{figure*}
\section{Method}
In this section, we will introduce an efficient 3D content generation framework for the controllable text-to-3D task, as shown in the Figure \ref{fig:framework}. Our framework not only accepts minimal text input but also allows for the specification of additional conditions, such as style, edges, scribbles, poses, or combinations of multiple conditions, enabling selective control based on individual preferences. Initially, based on the user's input, we employ a self-optimization process to generate high-quality, high-fidelity images (Section 3.1). We then review the 3D representation method we used, 3D Gaussian Splatting \cite{kerbl20233d} and apply 3D Gaussian distributions to transform the task into a generation process efficiently initialized through SDS \cite{poole2022dreamfusion} (Section 3.2). Meanwhile, we incorporate multi-view information, including multi-view depth, masks, features, and images, to constrain the generation process and refine textures (Section 3.3).

\subsection{Selectively Controllable Text to Image with Self-Optimization}
Images can convey richer and more precise visual information compared to text. Subtle details such as texture, color, and spatial relationships can be directly and explicitly captured in an image, which helps in generating more accurate and detailed 3D models.

Traditional text-to-image (T2I) models, such as Stable Diffusion \cite{rombach2022high}, have made significant progress in image generation, but they still have several shortcomings. First, these models often exhibit biases in semantic understanding, making it difficult to accurately capture textual descriptions and potentially leading to generated images that do not align with the user's expected results. Second, traditional models have weaker control over details, leading to images that lack precision, especially when it comes to intricate details. Moreover, high-quality results often depend on rich and detailed text prompts, yet in practical applications, users typically provide minimal text input while expecting high-quality visual outcomes. However, traditional models typically rely on static text input, and their ability to self-optimize and adjust is limited, making it difficult for them to iteratively improve. Lastly, when confronted with new or uncommon themes, traditional models may suffer from poor generalization, generating images that lack creativity and diversity. For example, when users desire artistic creations, they might want to input not only text but also additional condition images, which traditional models are unable to handle. Therefore, while these models possess strong generative capabilities, they still face challenges in semantic accuracy, detail handling, adaptability and creativity. At the same time, the quality of the generated images directly impacts the results of 3D generation, making high-quality image generation the foundation and key to successful 3D creation.

To facilitate artistic creation for users, our framework not only allows simple text input but also supports additional inputs such as style, edges, scribbles, poses, or combinations of several conditions. These inputs are then fed into a multi-modal iterative self-optimization framework for automated image design and generation. Following the approach of \cite{brade2023promptify, yang2023idea2img, hao2023optimizing, wang2024promptcharm}, our framework involves a Large Language Model (LLM), GPT-4V(ision) \cite{ai2023chatgpt, openai2023gpt, openai20234v, openai20234v}, interacting with T2I models to probe their usage and find effective T2I prompts. The LLM will play different roles in analyzing the feedback signals (i.e., candidate images) from the T2I model and designing the next round of queries (i.e., text-based T2I prompts). The three key processes of generating T2I prompts, selecting candidate images, and reflecting on feedback together enable the system's multimodal iterative self-optimization capability. Starting from the rough input text provided by the user, we first generate prompts using GPT-4V, which creates a set of N text prompts that align with the user's input, considering prior textual feedback and the history of optimization. In each iteration, GPT-4V carefully evaluates the N generated candidate images based on the same user input, assessing the quality of the generated images according to factors such as object count, attributes, entities, relationships, size, appearance, and overall similarity to the original user-provided prompt. A score between 0 and 10 is assigned to each candidate image, and the one with the most potential is selected. GPT-4V analyzes the differences between the candidate images and the original user input, providing insights into what is incorrect, possible reasons for the discrepancy, and suggestions for adjusting the T2I prompts to improve the image. After several iterations, we collect the image with the highest visual quality score, as judged by GPT-4V, to serve as the final base image for 3D content generation. Thus, our 3D content generation framework offers a user-friendly self-optimization process, eliminating the need for cumbersome prompt engineering like previous methods \cite{chung2023luciddreamer, wang2024prolificdreamer, poole2022dreamfusion}, while producing high-quality, visually appealing, and text-aligned images, which can later be converted into 3D content through Gaussian splatting.

\subsection{Generative Gaussian Splatting}
3D Gaussian Splatting is the latest breakthrough method in novel view synthesis, rendering images in real-time through Splatting. The DreamGaussian \cite{tang2023dreamgaussian} work first discovered that 3D Gaussian functions are highly effective for 3D generation tasks. 3D Gaussian Splatting represents 3D information using a set of 3D Gaussians. It leverages a set of 3D Gaussian primitives \( \theta \), view poses \( P \), and camera parameters containing the center \( o \) to compute per-pixel colors \( C \). Specifically, a Gaussian primitive can be described by its center \( \mu \in \mathbb{R}^3 \), scaling factor \( s \in \mathbb{R}^3 \), and rotation quaternion \( q \in \mathbb{R}^4 \). The \( i \)-th Gaussian primitive \( G_i \) can be represented as:
\begin{equation}
  G_i(\mathbf{x}) = e^{-\frac{1}{2} (\mathbf{x} - \boldsymbol{\mu}_i)^\top \boldsymbol{\Sigma}_i^{-1} (\mathbf{x} - \boldsymbol{\mu}_i)}
\end{equation}
The covariance matrix \( \boldsymbol{\Sigma} \) can be calculated based on the scaling \( \mathbf{s} \) and the rotation \( \mathbf{q} \) as \( \boldsymbol{\Sigma} = \mathbf{q} \mathbf{s} \mathbf{s}^\top \mathbf{q}^\top \). For rendering purposes, the Gaussian primitive also retains an opacity value \( \alpha \in \mathbb{R}^1 \) and a color \( \mathbf{c} \in \mathbb{R}^3 \). \( \theta_{i}=\{\mu_{i},s_{i},q_{i},\alpha_{i},c_{i}\} \) represents the parameters of the \( i \)-th Gaussian, which can also be written as \( \theta_i = \{\boldsymbol{\mu}_i, \boldsymbol{\Sigma}_i, \alpha_i, \mathbf{c}_i\} \). 
The 3D Gaussian Splatting represents a scene using a set of anisotropic Gaussians, defined by the center position \(\mu \in \mathbb{R}^3\), covariance \(\Sigma \in \mathbb{R}^7\), color \(c \in \mathbb{R}^3\), and opacity \(\alpha \in \mathbb{R}^1\). 

We initialize 3D Gaussians at random positions sampled inside a sphere with unit scaling and no rotation. These 3D Gaussian functions are regularly densified during optimization. Unlike reconstruction pipelines, we start with fewer Gaussians but densify them more frequently to stay aligned with the generation progress. We follow the recommendations from previous work \cite{poole2022dreamfusion, huang2023dreamtime, lin2023magic3d} and use SDS to optimize the 3D Gaussians. A different 2D diffusion prior $\phi$, is used to guide the SDS denoising step, which is backpropagated to the 3D Gaussians. DreamFusion \cite{poole2022dreamfusion} is one of the most representative methods for elevating 2D diffusion models to 3D. It proposes optimizing 3D representations using a pre-trained 2D diffusion model \(\phi\) through SDS loss. Specifically, it adopts MipNeRF \cite{barron2021mip} as the 3D representation method and optimizes its parameters \(\theta\). Assuming the rendering function is \(g\), the rendered image is \(x = g(\theta)\). To align the rendered image \(x\) with samples obtained from the diffusion model \(\phi\), DreamFusion uses a score estimation function: \(\hat{\epsilon}_{\phi}(z_t; y, t)\), which predicts the sampled noise \(\hat{\epsilon}_{\phi}\) based on the noisy image \(z_t\), text embedding \(y\), and noise level \(t\). By measuring the difference between the Gaussian noise \(\epsilon\) added to the rendered image \(x\) and the predicted noise \(\hat{\epsilon}_{\phi}\), this score estimation function provides a direction for updating the parameters \(\theta\). The gradient computation formula is:
\begin{equation}
\nabla_\theta \mathcal{L}_{\mathrm{SDS}}(\phi, \mathbf{x}=g(\theta)) \triangleq \mathbb{E}_{t, \epsilon}\left[w(t)\left(\hat{\epsilon}_\phi\left(\mathbf{z}_t ; y, t\right)-\epsilon\right) \frac{\partial \mathbf{x}}{\partial \theta}\right]
\end{equation}
where \(w(t)\) is a weighting function.

We integrate the multi-view diffusion models (MVDiffusion) \cite{shi2023mvdream, wang2023imagedream} model into our pipeline, using SDS based on four generated views. Specifically, in each iteration, four orthogonal views are rendered from a 3D Gaussian \( g(\phi) \) with random camera extrinsics and intrinsics \( \mathbf{c} \) for the four views. These views are then encoded into latent representations \( \boldsymbol{x}^{m v} \) and passed through the multi-view diffusion network to calculate the diffusion loss in the image space. The computed loss is backpropagated to optimize the parameters of the 3D Gaussian model. Formally, the multi-view SDS loss can be expressed as:
\begin{equation}
\mathcal{L}_{\mathrm{mv\_sds}}\left(\phi, \boldsymbol{x}^{m v}\right)=\mathbb{E}_{t, \mathbf{c}, \epsilon}\left[\left\|\boldsymbol{x}^{m v}-\hat{\boldsymbol{x}}_0^{m v}\right\|_2^2\right]
\end{equation}
where \( \hat{\boldsymbol{x}}_0^{mv} \) is the denoised multi-view image at timestep 0 from multi-view diffusion. After the above operations, it significantly improves the geometric accuracy of the generated objects, effectively eliminating the Janus problem.

We observed that even with multi-view SDS training iterations, the generated Gaussian functions often appear geometrically blurry and lack detail. This can be explained by the blurriness of the SDS loss. This observation prompted us to design the following further multi-view information constraints.

\subsection{Integrating Multi-View Information}
Generating 3D content based solely on a single image tends to result in poor quality, as it lacks information from other viewpoints, leading to issues like the multi-face Janus problem, especially when trying to generate the object's back. We integrate MVDiffusion and generate images from multiple viewpoints, such as four views from the front, back, left, and right, or other fixed perspectives. Generating 3D content based on multiple viewpoints significantly improves the result.

\subsubsection{Geometric Constraints}
The 3D Gaussian distribution surpasses neural radiance field methods in novel view synthesis by achieving lower computational costs and real-time high-quality rendering. Although it can generate high-quality renderings with a large number of input views, its performance significantly degrades when only a few views are available. To solve the above problems, we use monocular depth prediction as a prior and combine it with a scale-invariant depth loss to constrain the 3D shape under limited input views.

By specifying the camera's elevation, azimuth, and other parameters, we can obtain the corresponding rendered image and depth from the camera viewpoint, which allows us to compute the multi-view loss. In the original 3D Gaussian Splatting method, pixel-wise photometric loss is used to train the model. However, when there are very few views, this photometric loss alone is insufficient, as it tends to overfit the training views, and the 3D shape remains inaccurate. To address this, we propose using multi-view depth priors to constrain the 3D shape.

{\bfseries Gaussian Splatting Depth Rendering.} The 3D Gaussian Splatting represents a scene using a set of anisotropic Gaussians, defined by the center position \(\mu \in \mathbb{R}^3\), covariance \(\Sigma \in \mathbb{R}^7\), color \(c \in \mathbb{R}^3\), and opacity \(\alpha \in \mathbb{R}^1\). 3D Gaussian Splatting uses point-based differentiable rendering \cite{kopanas2021point, kopanas2022neural} and computes the color \( C \) of a pixel by blending \(N\) ordered Gaussians that overlap with the pixel.
\begin{equation}
C=\sum_{i \in N} c_i \alpha_i \prod_{j=1}^{i-1}\left(1-\alpha_j\right)
\end{equation}
Similar to pixel rendering, depth rendering uses alpha blending from the z-buffer of ordered Gaussians. The depth \( D \) of a point is computed as follows, where \( d_i \) is the z-buffer value of the \( i \)-th Gaussian.
\begin{equation}
D=\sum_{i \in N} d_i \alpha_i \prod_{j=1}^{i-1}\left(1-\alpha_j\right)
\end{equation}

{\bfseries Scale Invariant Depth Loss.} For a given set of viewpoints, we calculate the depth estimated by the monocular depth estimation model and the depth rendered from the corresponding viewpoint, respectively. We can then compute the loss between the two for supervision. One of the challenges in computing the depth loss is the scale of the depth values. The scale between the monocular model and the 3D Gaussian Splatting depth renderer may differ. Therefore, based on the work of Eigen et al. \cite{eigen2014depth}, we use a scale-invariant loss function that takes into account the scales of both the monocular and 3D Gaussian Splatting depth renderers. Given the rendered depth map \( D \) and the predicted depth \( \tilde{D} \), each with \( n \) pixel indices \( i \), we compute the scale-invariant mean squared error as follows:
\begin{equation}
\mathcal{L}_{scale}\left(D, \tilde{D}\right) = \frac{1}{2n} \sum_{i=1}^n \left( \log D_{i} - \log \tilde{D}_{i} + \alpha\left(D_{i}, \tilde{D}_{i}\right) \right)^2
\end{equation}
where \( \alpha(D, \tilde{D}) = \frac{1}{n} \sum_i \left( \log \tilde{D}_i - \log D_i \right) \) is the value of \( \alpha \) that minimizes the error for a given pair \( (D, \tilde{D}) \).

To ensure the robustness of depth prediction across different scales, we calculate the multiscale loss \( \mathcal{L}_{multiscale} \) by interpolating the predicted depth and ground truth depth maps at various resolutions. Specifically, we use different scaling factors to scale both the predicted and ground truth depth maps, and compute the L1 loss at each scale. The multiscale loss is then obtained by summing these individual losses with a small scaling factor to prevent large gradient updates.
\begin{equation}
\mathcal{L}_{multiscale} = \sum_{s \in scales} \lambda_s \cdot \| D_s - \tilde{D}_s \|_1
\end{equation}
This approach enables the model to capture finer details at different resolutions, significantly improving the overall depth estimation performance, especially in regions with sparse or unclear depth information. 
In addition to the multiscale loss, we also used \( \mathcal{L}_{huber} \), which provides stronger smoothness and faster gradient convergence while reducing sensitivity to outliers. Compared to L2 loss, Huber loss is less sensitive to large errors, making it more robust in depth prediction, especially when the ground truth depth contains noise or inconsistencies. Overall, our depth loss function is represented as follows, where \(\alpha\), \(\beta\), and \(\gamma\) are the weight coefficients for the corresponding loss functions.
\begin{equation}
\mathcal{L}_{depth} = \alpha \mathcal{L}_{scale} + \beta \mathcal{L}_{multiscale} + \gamma \mathcal{L}_{huber}
\end{equation}
By combining these depth loss functions, our model can provide better geometric constraints. In addition, we can also add mask loss to further impose geometric constraints. In our framework, multi-view images are first obtained through the MVDiffusion. The image masks \( \boldsymbol{\tilde{I}}^{m v} \) are then extracted from the alpha channels of these images. The loss is subsequently calculated by comparing these masks with the transparency obtained through Gaussian rendering \( \boldsymbol{I}^{m v} \).
\begin{equation}
\mathcal{L}_{mask} = \|\boldsymbol{I}^{m v} - {\boldsymbol{\tilde{I}}^{m v}}\|_2^2
\end{equation}

\subsubsection{Feature Loss}
Previous studies have emphasized the importance of capturing the appearance of objects to establish robust feature correspondences between different views \cite{henaff2022object, rother2018g}. Inspired by these studies, our goal is to establish a visual feature correspondence between the reference view and the rendered view. This is achieved by enforcing feature-level similarity, guiding 3D Gaussians to effectively fill the geometric gaps in the invisible regions.

We utilize the DINOv2 features \cite{oquab2023dinov2}, which are more robust to appearance variations, and construct our loss function based on the cosine similarity between the DINOv2 features of the input image and the rendered image. Since DINOv2 features are defined at the image patch level rather than at the pixel level, they enable optimization of high-level semantic consistency in the images. In image generation tasks, even though traditional pixel-level losses (such as L1 loss) can reduce low-level pixel differences, they may overlook high-level semantic information in the image, such as object shapes, colors, and structures. By incorporating feature loss, we can ensure that the generated image is closer to the reference image in terms of semantic content, which is essential for producing more natural and consistent images. The feature loss can be defined as:
\begin{equation}
\mathcal{L}_{feature} = 1 - \varphi(\mathbf{f}, \tilde{\mathbf{f}}) \quad \varphi(\mathbf{f}, \tilde{\mathbf{f}}) = \frac{\mathbf{f} \cdot \tilde{\mathbf{f}}}{\|\mathbf{f}\|_2 \|\tilde{\mathbf{f}}\|_2}
\end{equation}
where \( \mathbf{f} \) and \( \mathbf{\tilde{f}} \) are the feature representations of the render and reference images, respectively. \( \varphi(\mathbf{f}, \tilde{\mathbf{f}}) \) is the cosine similarity between the features of the render image and the reference image.

\subsubsection{Texture Refinement}
Polygonal mesh is a widely used 3D representation, especially in industrial applications. We refer to the mesh extraction method in \cite{tang2023dreamgaussian}, which is primarily based on the Marching Cubes algorithm. After acquiring the mesh geometry, we can project the rendered RGB image back onto the mesh surface and use it as a texture. To do this, we begin by unwrapping the mesh's UV coordinates and initializing an empty texture map. We then select some azimuth and elevation angles, in addition to the top and bottom views, to render the corresponding RGB images. Each pixel from these rendered images is back-projected onto the texture map using the UV coordinates. In line with \cite{richardson2023texture}, we exclude pixels with small z-direction camera-space normals to prevent unstable projections at the mesh boundaries. This back-projected texture map serves as the starting point for the subsequent texture refinement stage on the mesh.

With the initialization texture in hand, we first render a blurry image \( \mathcal{I^*}\) from an arbitrary camera view. Next, we introduce random noise into the image and apply a multi-step denoising process \( f_\phi(\cdot) \) using a 2D diffusion prior to obtain a refined image:
\begin{equation}
\mathcal{I}=f_\phi\left(\mathcal{I^*}+\epsilon\left(t_{\mathrm {start}}\right) ; t_{\mathrm {start}} \right)
\end{equation}
Specifically, let \( \epsilon(t_{\mathrm{start}}) \) denote the random noise at the starting timestep \( t_{\mathrm{start}} \), which is carefully selected to control the noise strength. This ensures that the refinement enhances details without distorting the original content. The resulting refined image is then used to optimize the texture via a pixel-wise mean squared error loss:
\begin{equation}
\mathcal{L}_{refine}=\left\|\mathcal{I}-\mathcal{I^*}\right\|_2^2
\end{equation}

A well-known zero-shot method is SDS \cite{poole2022dreamfusion}, which has shown particular effectiveness in 3D generation and texture refinement \cite{lin2023magic3d, wang2024prolificdreamer, youwang2024paint, metzer2023latent}. However, as observed, applying this diffusion model at this step leads to suboptimal results and requires high CFG \cite{ho2022classifier} weights to converge, which can cause oversaturation. Therefore, we adopt a method that simultaneously updates the specified four viewpoints to optimize the texture. We compute the loss between the rendered images from the four specified viewpoints and the corresponding images generated by MVDiffusion from the previous four viewpoints.
\begin{equation}
\mathcal{L}_{color} = (1 - \lambda) \mathcal{L}_1(\boldsymbol{I}^{m v}, \boldsymbol{\tilde{I}}^{m v}) + \lambda \mathcal{L}_{\text{D-SSIM}}(\boldsymbol{I}^{m v}, \boldsymbol{\tilde{I}}^{m v})
\end{equation}


\section{Experiment}
In this section, we begin by outlining the implementation details in Section 4.1, followed by a presentation of comparisons and visualization results in Section 4.2. Section 4.3 focuses on a set of ablation studies to assess the effectiveness of our method.

\subsection{Implementation Details}
We divide the training process into two stages: the first stage consists of 300 steps, and the second stage consists of 60 steps. If the geometric shapes of the content to be generated are relatively simple, the steps in the first stage can be appropriately reduced. The resolution of the rendered images increases from 128 to 512 according to a specific pattern: during the first 30\% of the training time, a resolution of 128 is used; during the 30\%-60\%, the resolution is 256; and in the remaining time, the resolution is 512. In the early stages of training, we randomly render viewpoints with the camera azimuth in the range of [-180, 180] degrees and the elevation in the range of [-30, 30] degrees. In the final 1/6 of the training, we place more emphasis on optimizing the four viewpoints at 0°, 90°, 180°, and -90° to enhance the consistency of the viewpoint results. At the same time, during the training process, we perform densification and pruning operations on the Gaussian points to optimize the point distribution in the view space. All experiments are performed and measured using a single NVIDIA 3090 (24GB) GPU.

\begin{figure*}[htbp]
    \centering
    \includegraphics[width=0.8\textwidth]{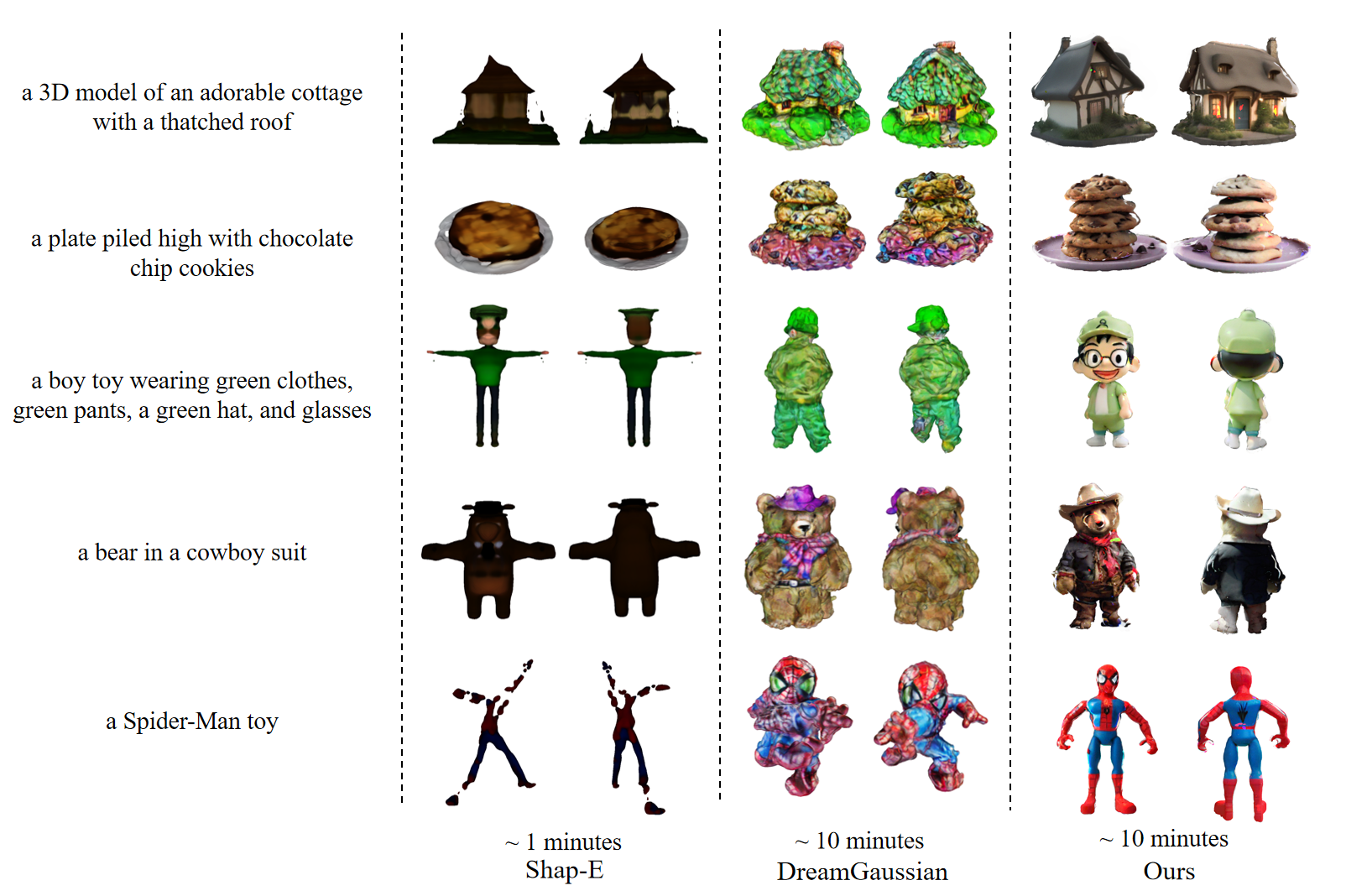}
    \caption{A comparison of our method with Shap-E and DreamGaussian. Shap-E, DreamGaussian, and our method can all generate 3D content in a short amount of time. Our method, with a time cost similar to DreamGaussian, is able to generate more complex and richer 3D content compared to Shap-E and DreamGaussian.}
    \Description{A comparison of our method with Shap-E and DreamGaussian.}
    \label{fig:compare1}
\end{figure*}

\begin{figure*}[htbp]
    \centering
    \includegraphics[width=0.9\textwidth]{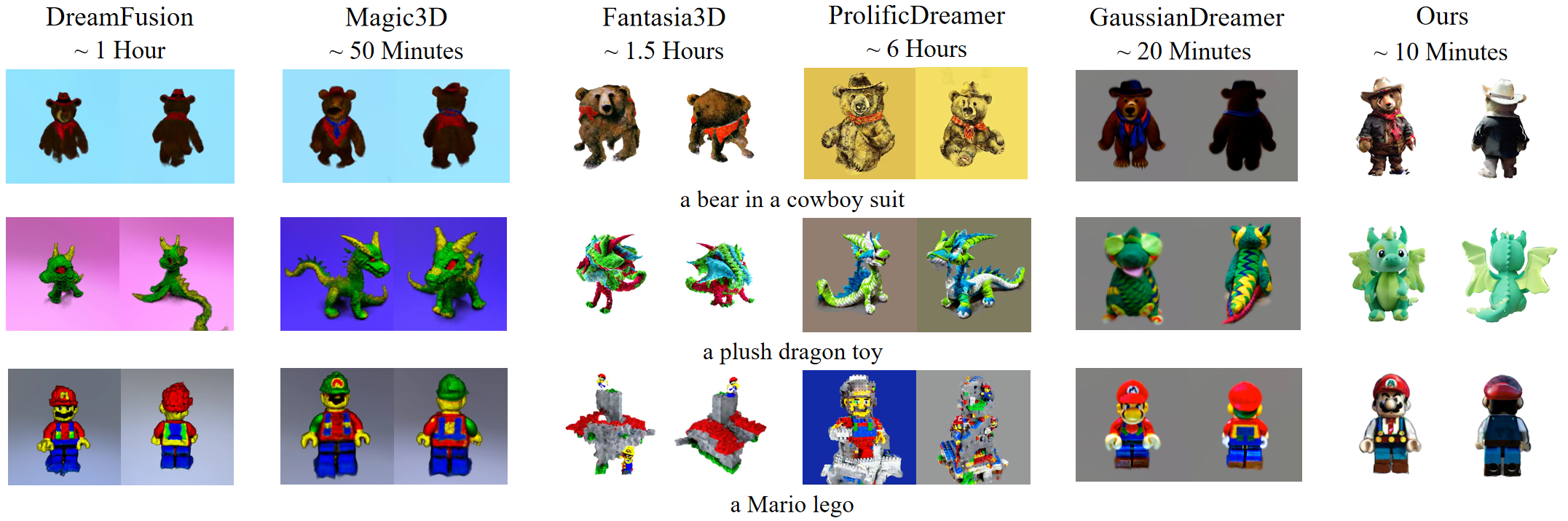}
    \caption{Comparisons on Text-to-3D methods. Among these methods, our approach takes the least time to generate 3D content while maintaining viewpoint consistency.}
    \Description{Comparisons on Text-to-3D methods.}
    \label{fig:compare2}
\end{figure*}

\subsection{Comparisons and Visualization Results}
We first present a comparison of our method with other text-to-3D methods in Figures \ref{fig:compare1} and \ref{fig:compare2}. 
In Figure 4, for each method, we present both the front view (first column) and the back view (second column) of the generated results. DreamFusion, Magic3D, Fantasia3D, and ProlificDreamer all utilize NeRF as a 3D representation, which incurs higher time costs and suffers from the Janus problem. The outputs from DreamFusion, Magic3D, and ProlificDreamer for "a plush dragon toy" exhibit a "head" in both the front and back views, with DreamFusion and Magic3D additionally generating two tails.
For all the compared methods, our approach demonstrates better generation quality and shorter generation time. Our method achieves a better balance between generation quality and speed. 

\begin{figure}[htbp]
    \centering
    \includegraphics[width=0.7\textwidth]{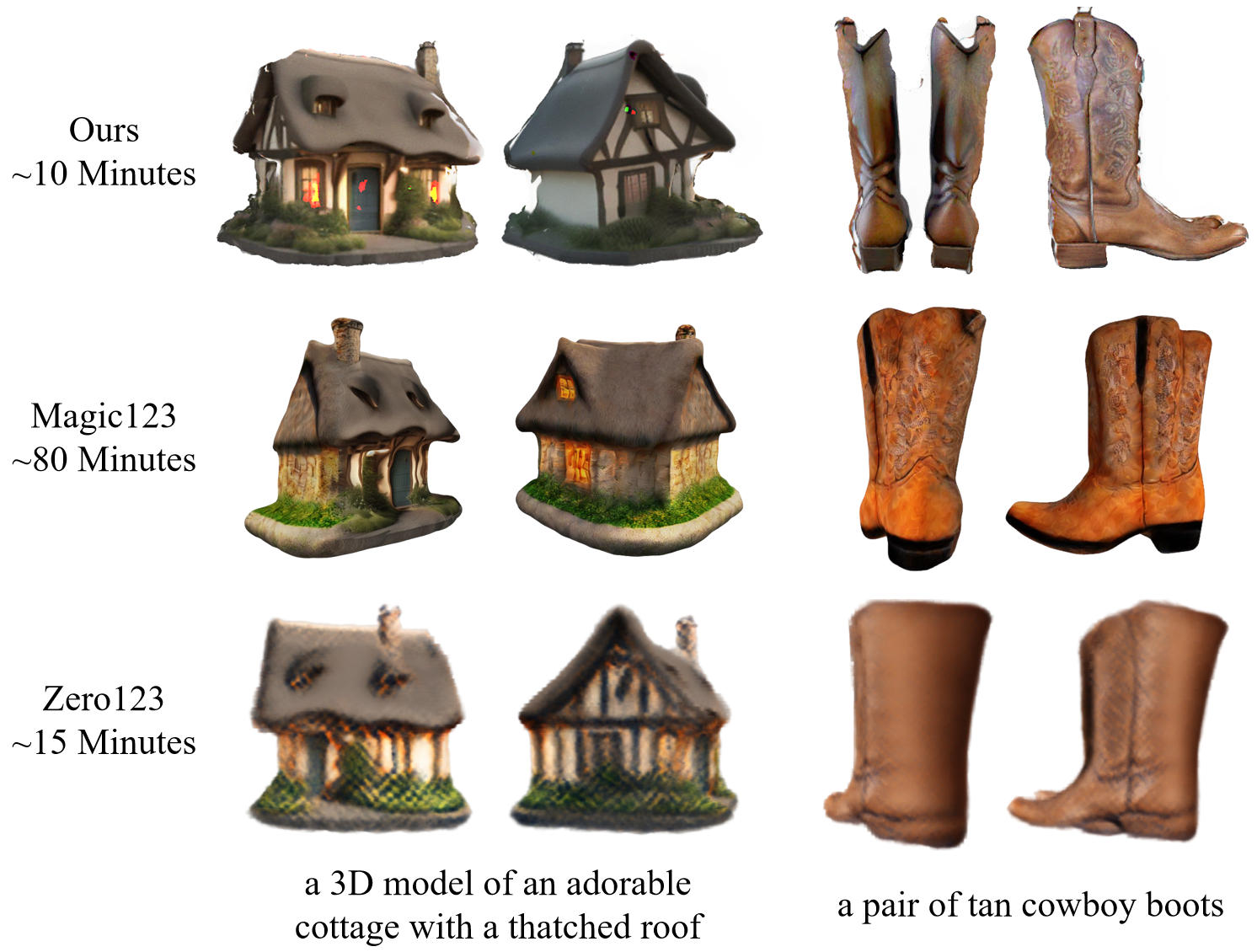}
    \caption{Comparisons on Image-to-3D methods. Compared to Zero123 and Magic123, our method generates more reasonable and consistent content in a shorter time.}
    \label{fig:compare4}
\end{figure}

In Figure \ref{fig:compare4}, we further compare our method with other Image-to-3D approaches. Our method requires less time for generation while ensuring multi-view consistency. Additionally, in the example of "a pair of tan cowboy boots," both Zero123 and Magic123 fail to generate a complete pair of boots, producing only a single boot instead. In contrast, our method successfully generates a full pair of boots.
Moreover, our method can extract texture meshes, making it highly suitable for integration into downstream applications. For instance, using software like Blender \cite{blender2018blender}, we can easily utilize these meshes for assembly and animation purposes.

In Table \ref{tab:clip_comparison}, we quantitatively evaluate our method using CLIP \cite{radford2021learning} similarity. We report the CLIP similarity and average generation time for different text-to-3D methods, calculating the similarity between the images generated by each method and the corresponding text. We use two models to compute the CLIP similarity: ViT-Large-patch14 from OpenAI \cite{radford2021learning} and CLIP-ViT-bigG-14-laion2B-39B-b160k from LAION \cite{zhai2019large}. Our method outperforms all the compared methods in terms of both generation quality and speed.

\begin{table}[h]
\centering
\caption{Comparisons on CLIP similarity with other text-to-3D methods.}
\label{tab:clip_comparison}
\begin{tabular}{l|ccc}
\toprule
\textbf{}                                                     & ViT-Large-patch14 $\uparrow$ & ViT-bigG-14 $\uparrow$ & Time $\downarrow$ \\ \midrule
Shap-E \cite{jun2023shap}                    & 24.53                        & 40.64                          & 1 minute                    \\
DreamFusion \cite{poole2022dreamfusion}      & 25.16                        & 45.14                          & 1 hour                       \\
Magic3D \cite{lin2023magic3d}                & 26.72                        & 46.09                          & 50 minutes                   \\
Fantasia3D \cite{chen2023fantasia3d}                                                   & 24.44                        & 36.99                          & 1.5 hours                    \\
ProlificDreamer \cite{wang2024prolificdreamer}                                              & 28.71                        & 48.62                          & 6 hours                      \\
DreamGaussian \cite{tang2023dreamgaussian}   & 28.68                        & 45.29                          & 10 minutes                   \\
GaussianDreamer \cite{yi2024gaussiandreamer} & 30.55                        & 49.42                          & 20 minutes                   \\
Ours                                                          & \textbf{31.16}                        & \textbf{49.90}                          & \textbf{10 minutes}                   \\ \bottomrule
\end{tabular}
\end{table}

For a comprehensive evaluation, we conducted an extensive numerical assessment using three metrics that cover different aspects of image similarity: SSIM \cite{wang2004image}, PSNR, and LPIPS \cite{zhang2018unreasonable}. These metrics are evaluated in the reference view to measure reconstruction quality and perceptual similarity. As shown in Table \ref{tab:img_compare}, our method achieves the best performance across all metrics when compared to Zero123 and Magic123.

\begin{table}[h]
\caption{Comparison of different image-to-3D methods.}
\label{tab:img_compare}
\begin{tabular}{c|ccc}
\toprule
      & Zero123 \cite{liu2023zero} & Magic123 \cite{qian2023magic123} & Ours   \\ \midrule
SSIM $\uparrow$  & 0.133   & 0.083    & \textbf{0.814}  \\
PSNR $\uparrow$  & 7.606   & 7.175    & \textbf{10.079} \\
LPIPS $\downarrow$ & 0.443   & 0.460    & \textbf{0.234}  \\ \bottomrule
\end{tabular}
\end{table}

\begin{figure}[h]
    \centering
    \includegraphics[width=\textwidth]{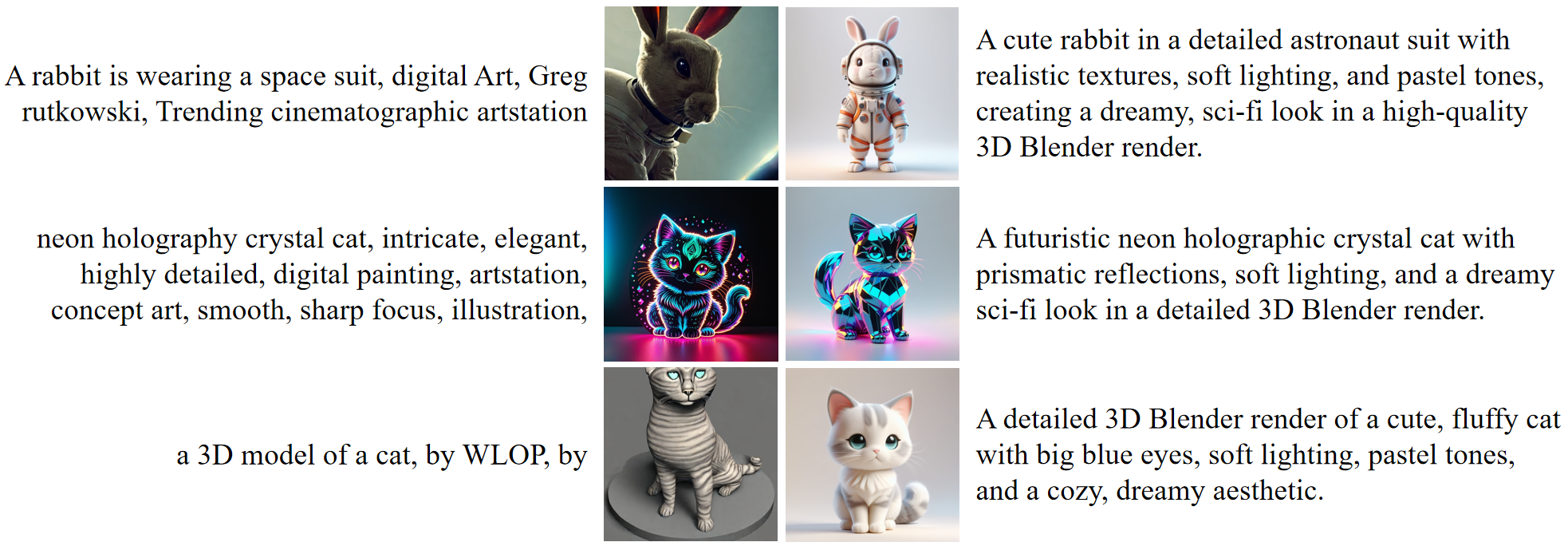}
    \caption{Promptist vs. Ours: Prompt Optimization results. The left side represents the Promptist results, while the right side represents our results.}
    \label{fig:compare_prompt}
\end{figure}

Figure \ref{fig:compare_prompt} presents a comparison between our prompt optimization results and those of Promptist \cite{hao2023optimizing}. Previous work \cite{hao2023optimizing, wang2024promptcharm} often produces prompts that, when used for image generation, fail to generate complete object structures, making the resulting images unsuitable as a foundation for 3D generation (as shown on the left side of the figure). Since our primary focus is on optimizing prompts for high-quality 3D content generation, our method places greater emphasis on aspects such as the style, geometry, texture, and pose of the objects, while also highlighting elements such as "soft lighting," "soft colors," and "3D Blender render." Consequently, our prompt optimization method is better suited for generating 3D content (as shown on the right side of the figure).

In addition, we compare the performance with the latest work MVControl \cite{li2024controllable}, in terms of controllable generation, as shown in the Figures \ref{fig:compare_Ironman} and \ref{fig:compare_tulip}. The Figures \ref{fig:compare_Ironman} shows that when we input the same text and scribble condition, the second row displays our generated results, which take 1/10 of the time compared to MVControl. The generated content is more complete, featuring a full mesh, and can effectively estimate content from other viewpoints. Additionally, details such as hands and feet are handled more accurately. Based on the comparison, it can be observed that our method generates more complete and detailed content in a shorter amount of time. Figure \ref{fig:compare_tulip} shows that when we input the same text and canny condition, the second row displays our results, where the visual effect of our multi-view prediction is significantly better than the first row (MVControl). Due to MVControl's poor performance in multi-view prediction, it directly affects the quality of its 3D generation, leading to irregular geometries. In contrast, our method ensures multi-view consistency, so the final generated content has more uniform and regular geometry.
\aptLtoX[graphic=no,type=html]{\begin{figure}[h]
    \centering
    \begin{minipage}{0.5\textwidth}
        \centering
        \includegraphics[width=\textwidth]{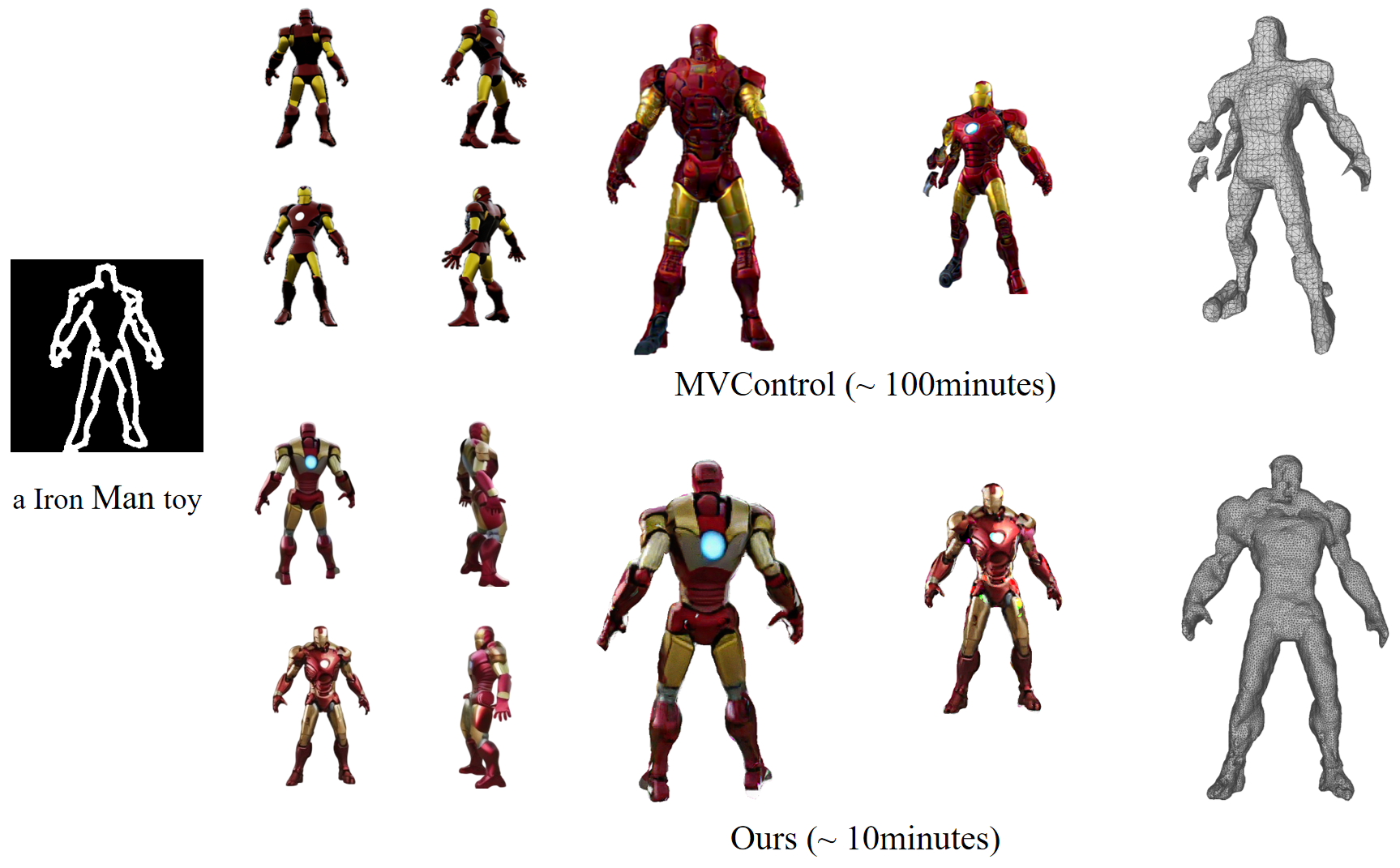}
        \caption{Ours vs. MVControl: Text and Scribble-Based.}
        \label{fig:compare_Ironman}
    \end{minipage}%
\end{figure}
\begin{figure}
    \begin{minipage}{0.5\textwidth}
        \centering
        \includegraphics[width=\textwidth]{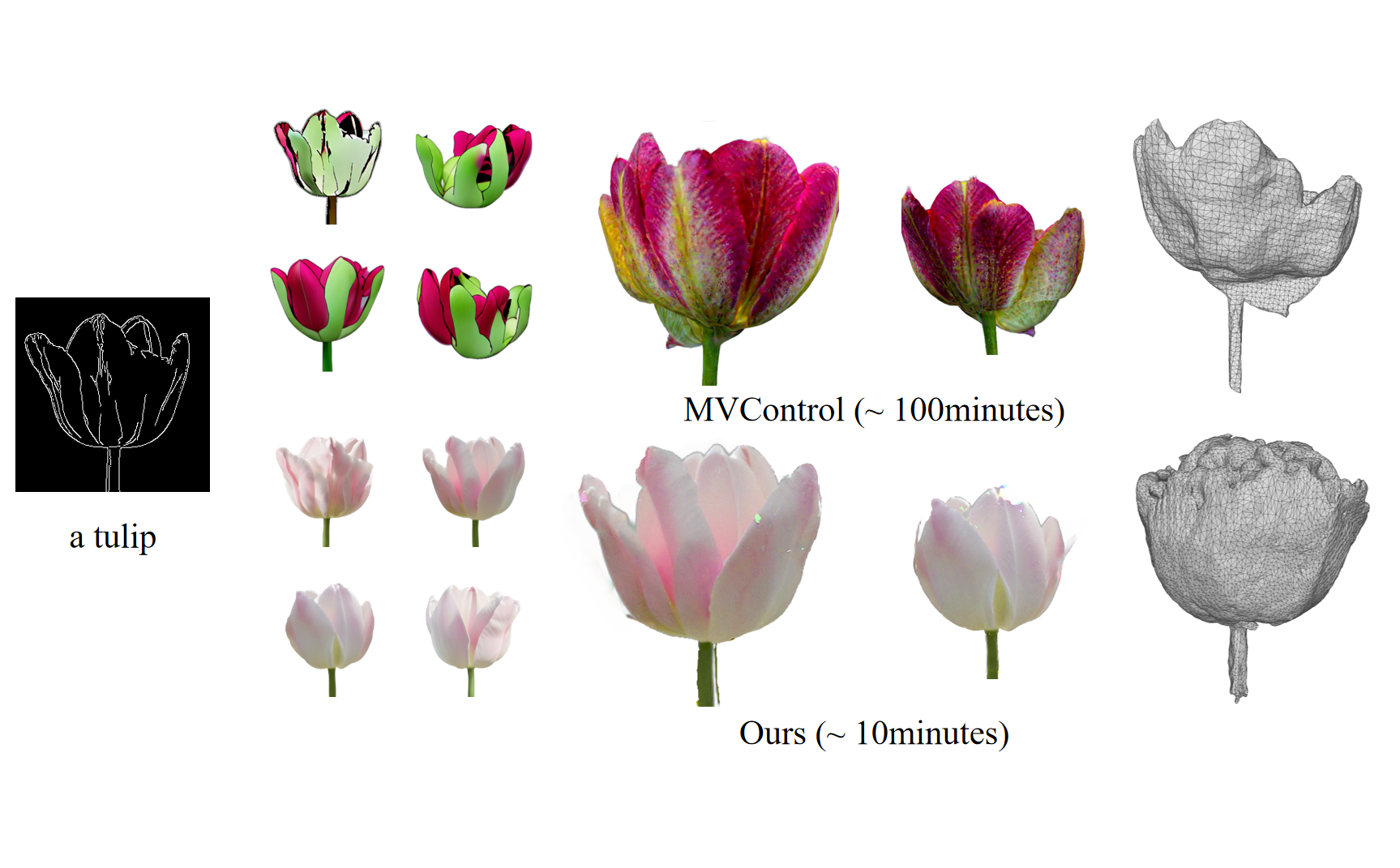}
        \caption{Ours vs. MVControl: Text and Canny-Based.}
        \label{fig:compare_tulip}
    \end{minipage}
\end{figure}}{\begin{figure}[h]
    \centering
    \begin{minipage}{0.5\textwidth}
        \centering
        \includegraphics[width=\textwidth]{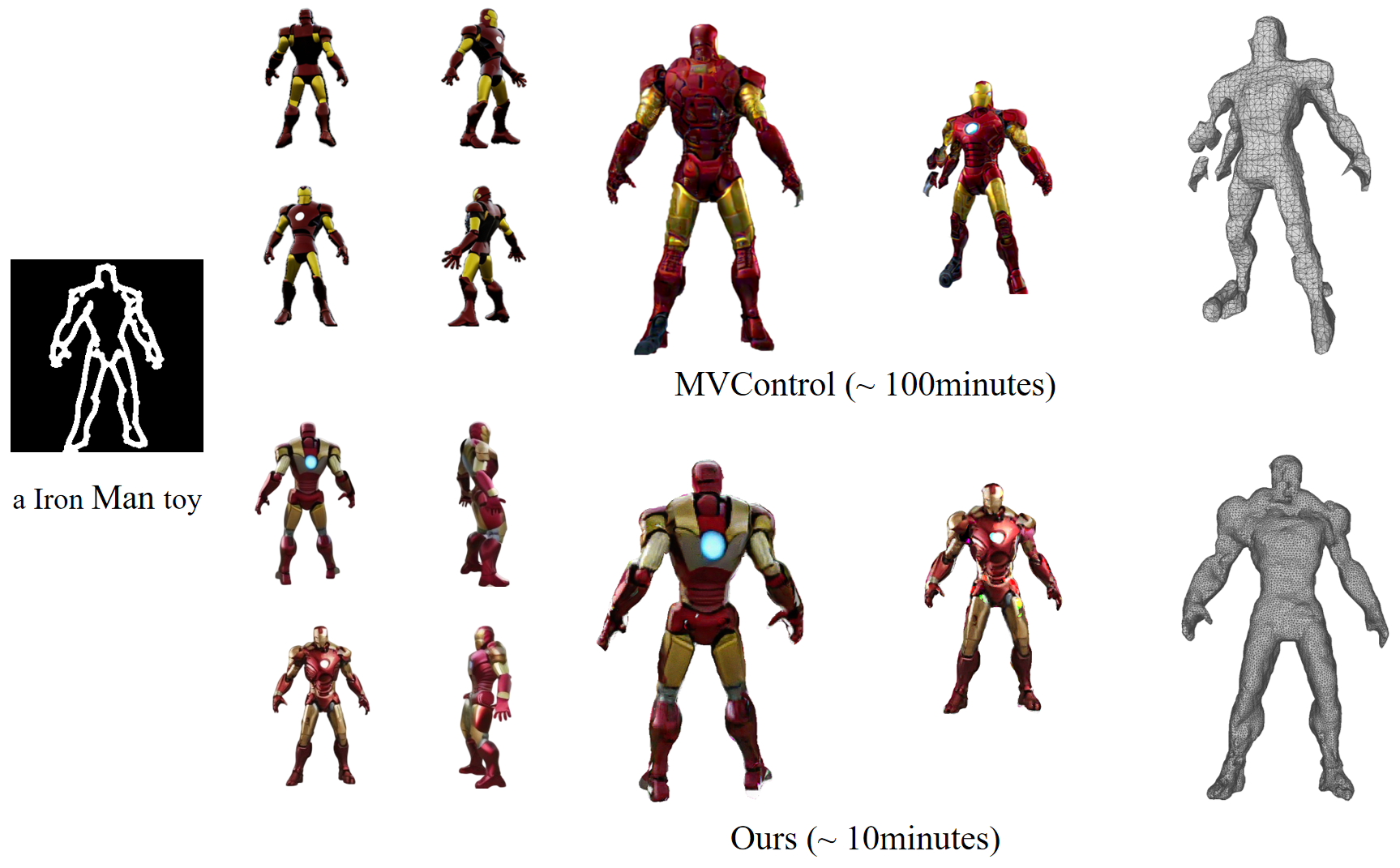}
        \caption{Ours vs. MVControl: Text and Scribble-Based.}
        \label{fig:compare_Ironman}
    \end{minipage}%
    \begin{minipage}{0.5\textwidth}
        \centering
        \includegraphics[width=\textwidth]{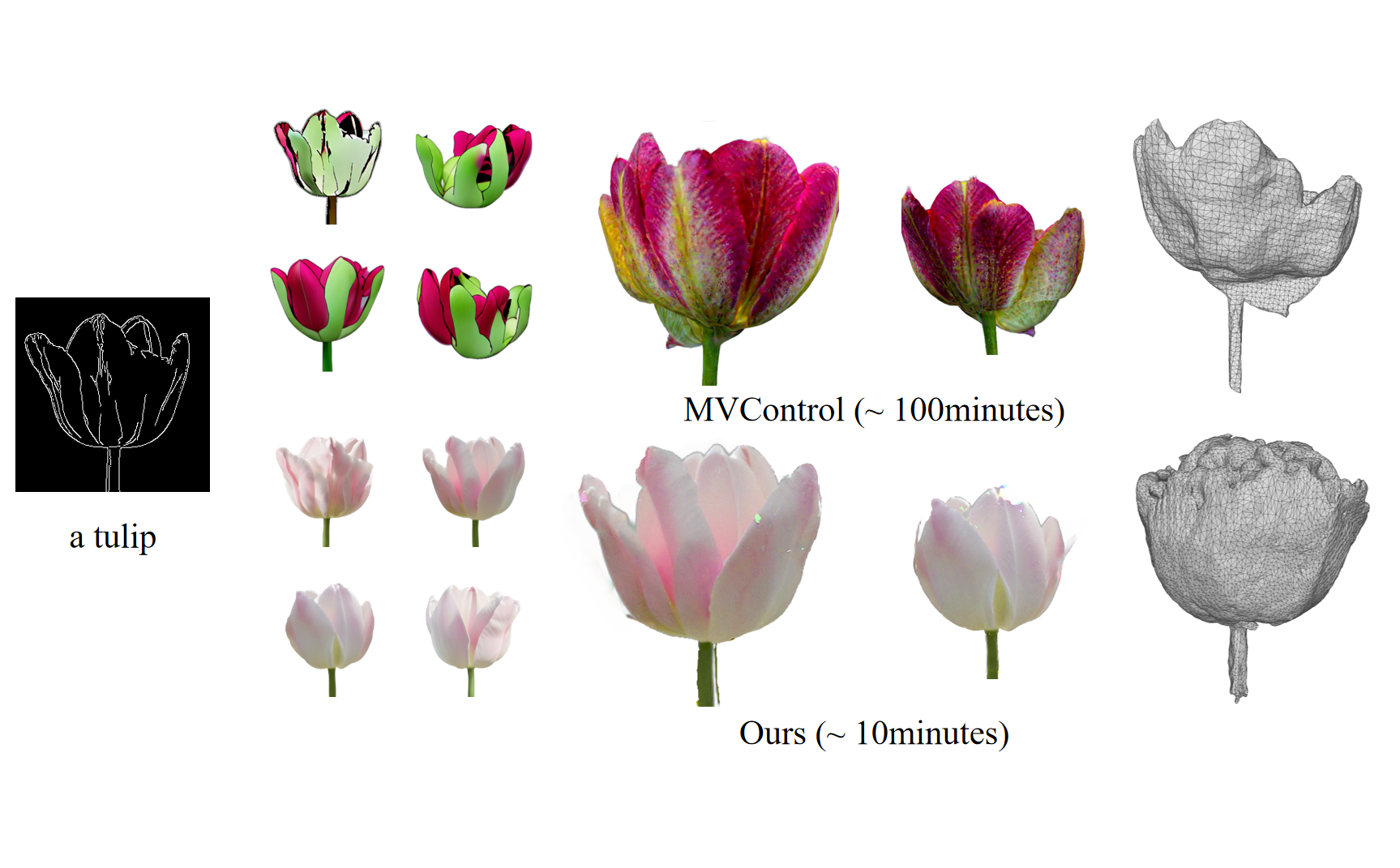}
        \caption{Ours vs. MVControl: Text and Canny-Based.}
        \label{fig:compare_tulip}
    \end{minipage}
\end{figure}}

We also demonstrate the capability of our method in artistic creation in Figures \ref{fig:idea_dog} and \ref{fig:idea_Ironman}. Figure \ref{fig:idea_dog} shows that our framework can extract the content of images, such as the color, shape, and style of a hat, and perform combinatorial artistic creation, generating harmonious content. In addition, Figure \ref{fig:idea_Ironman} also demonstrates that our framework can effectively understand the content of images, such as the character's appearance and physical features, can also produce coherent, high-quality 3D content. Our method has a strong ability to understand both text and images, enabling the creation of rich and diverse 3D content. By providing only a very brief text and the condition image to be input, controllable content that meets the desired expectations can be generated.
\aptLtoX[graphic=no,type=html]{\begin{figure}[htbp]
    \centering
    \begin{minipage}{0.5\textwidth}
        \centering
        \includegraphics[width=\textwidth]{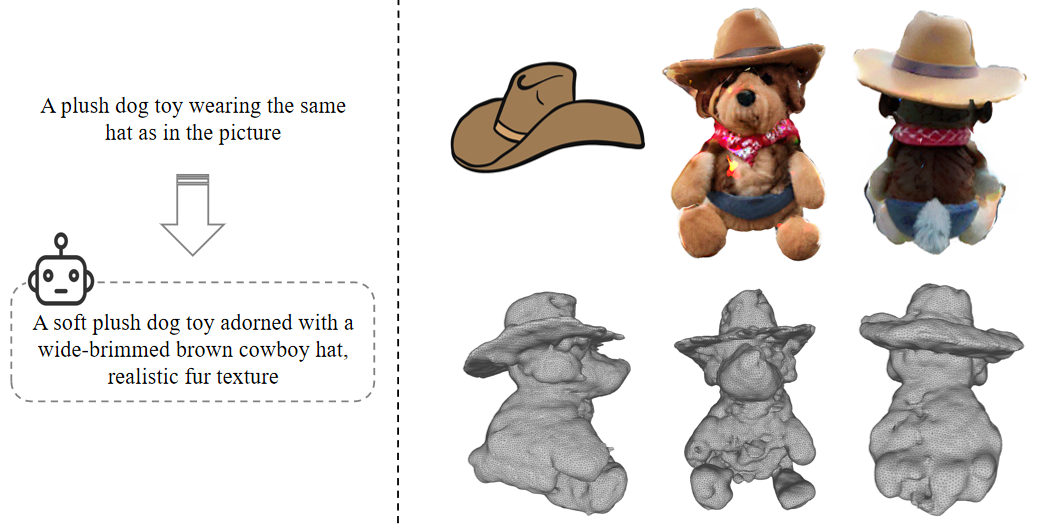}
        \caption{Combinatorial Art Creation.}
        \label{fig:idea_dog}
    \end{minipage}%
\end{figure}
\begin{figure}
    \begin{minipage}{0.5\textwidth}
        \centering
        \includegraphics[width=\textwidth]{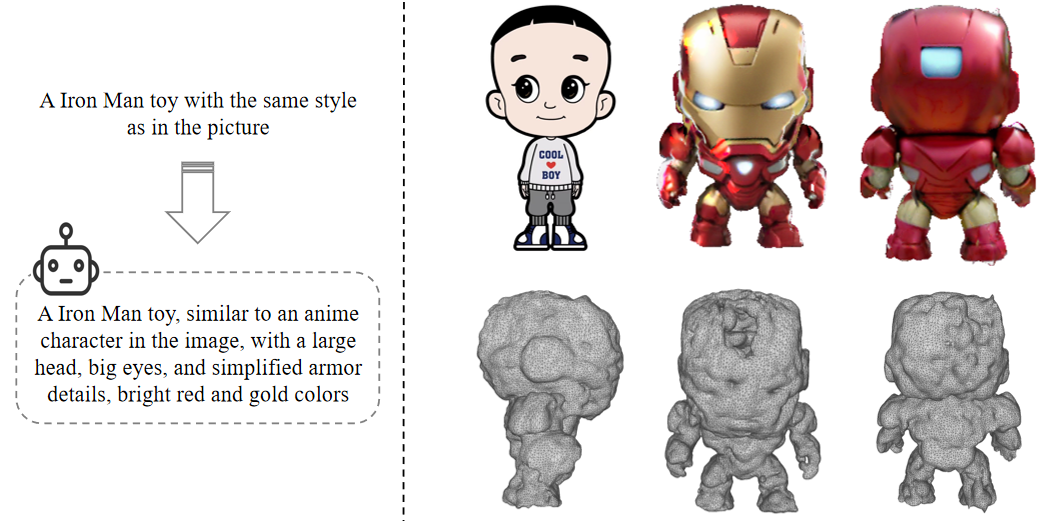}
        \caption{Understanding-Based Art Creation.}
        \label{fig:idea_Ironman}
    \end{minipage}
\end{figure}
}{
\begin{figure}[htbp]
    \centering
    \begin{minipage}{0.5\textwidth}
        \centering
        \includegraphics[width=\textwidth]{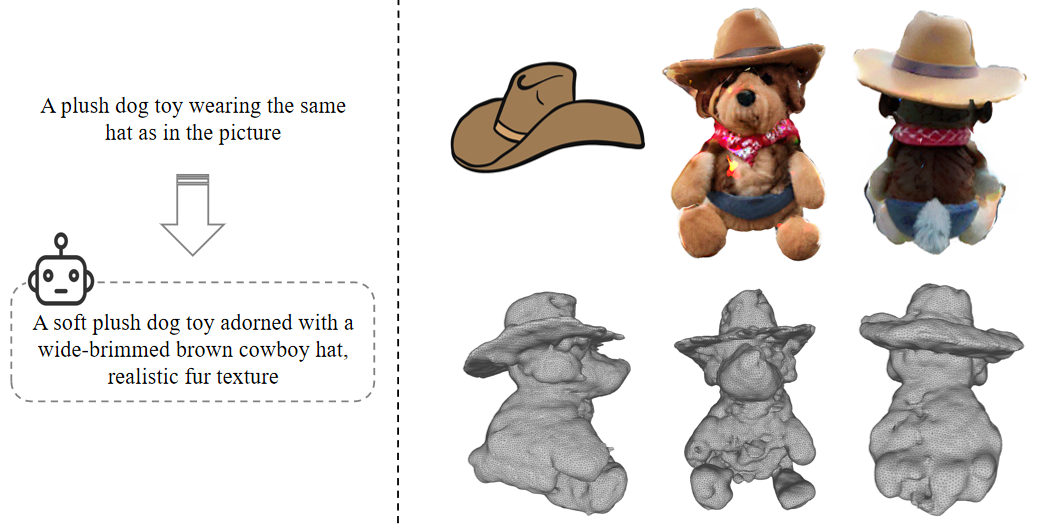}
        \caption{Combinatorial Art Creation.}
        \label{fig:idea_dog}
    \end{minipage}%
    \begin{minipage}{0.5\textwidth}
        \centering
        \includegraphics[width=\textwidth]{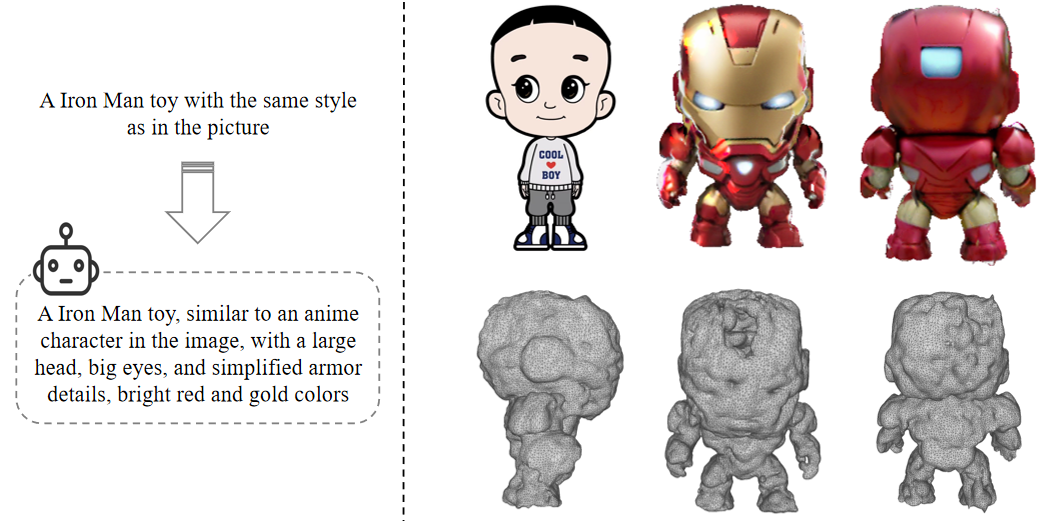}
        \caption{Understanding-Based Art Creation.}
        \label{fig:idea_Ironman}
    \end{minipage}
\end{figure}
}


\subsection{Ablation Study and Analysis}
We present the ablation study of multiscale depth in Figure \ref{fig:ablation_depth}. The first row shows the results without multiscale depth, while the second row demonstrates the results with multiscale depth. As observed, incorporating multiscale depth significantly improves overall depth estimation performance, particularly in regions with sparse or ambiguous depth information, such as object boundaries, hands, feet, and head areas. The Figure \ref{fig:ablation_mv} illustrates the ablation results of multi-view information, including depth, masks, features, and other relevant information. The first row shows the rendering results with multi-view information, while the second row shows the rendering results without multi-view information. It can be observed that without the multi-view information constraint, the rendering results from perspectives other than the frontal view are significantly worse, lacking in detail. The ablation study demonstrates that our method effectively improves the generation quality from non-frontal perspectives, significantly enhancing the overall 3D content generation quality.

\aptLtoX[graphic=no,type=html]{\begin{figure}[h]
    \centering
    \begin{minipage}{0.6\textwidth}
        \centering
        \includegraphics[width=\textwidth]{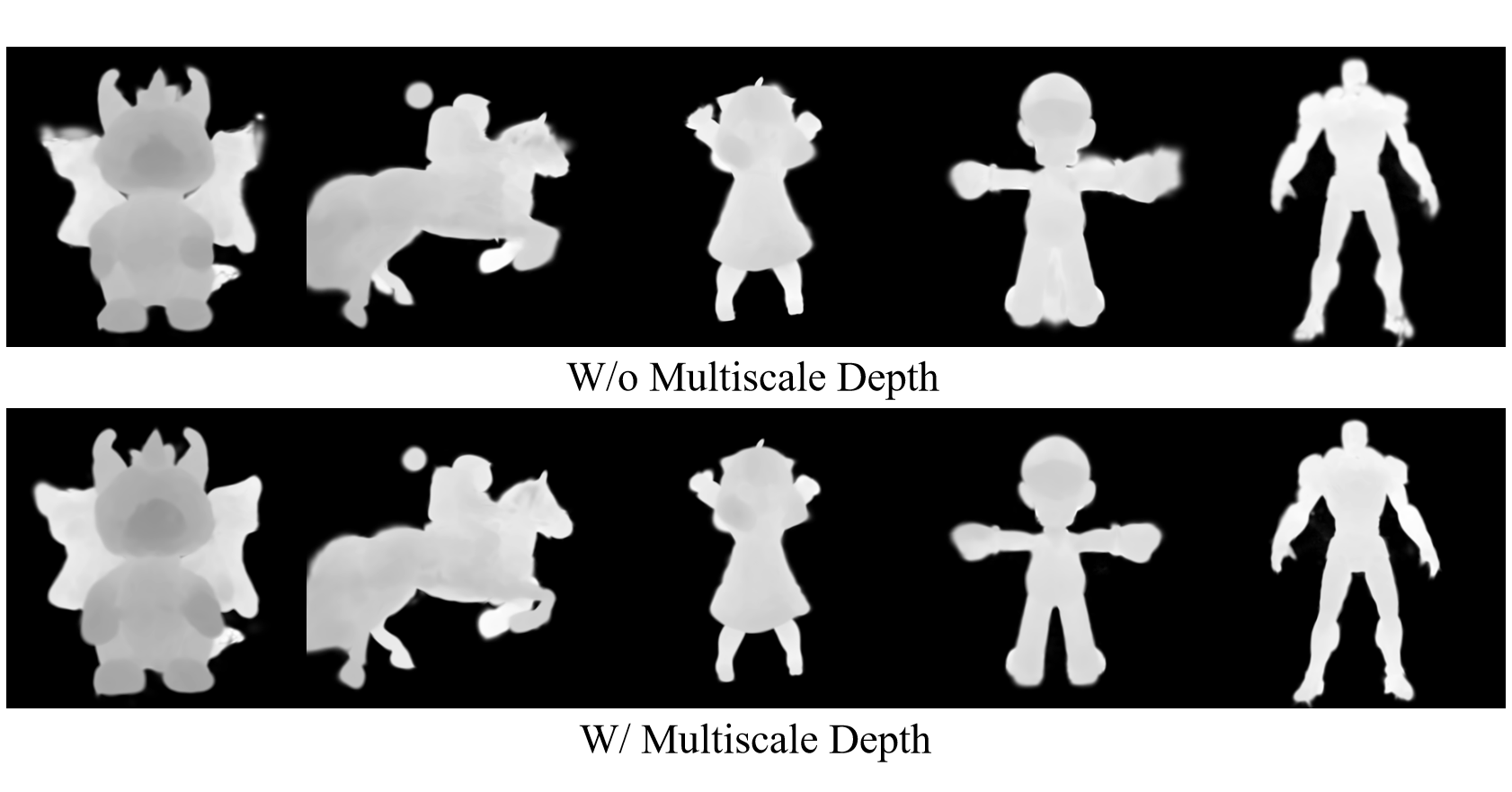}
        \caption{Ablation of Multiscale Depth.}
        \label{fig:ablation_depth}
    \end{minipage}%
\end{figure}
\begin{figure}
    \begin{minipage}{0.4\textwidth}
        \centering
        \includegraphics[width=\textwidth]{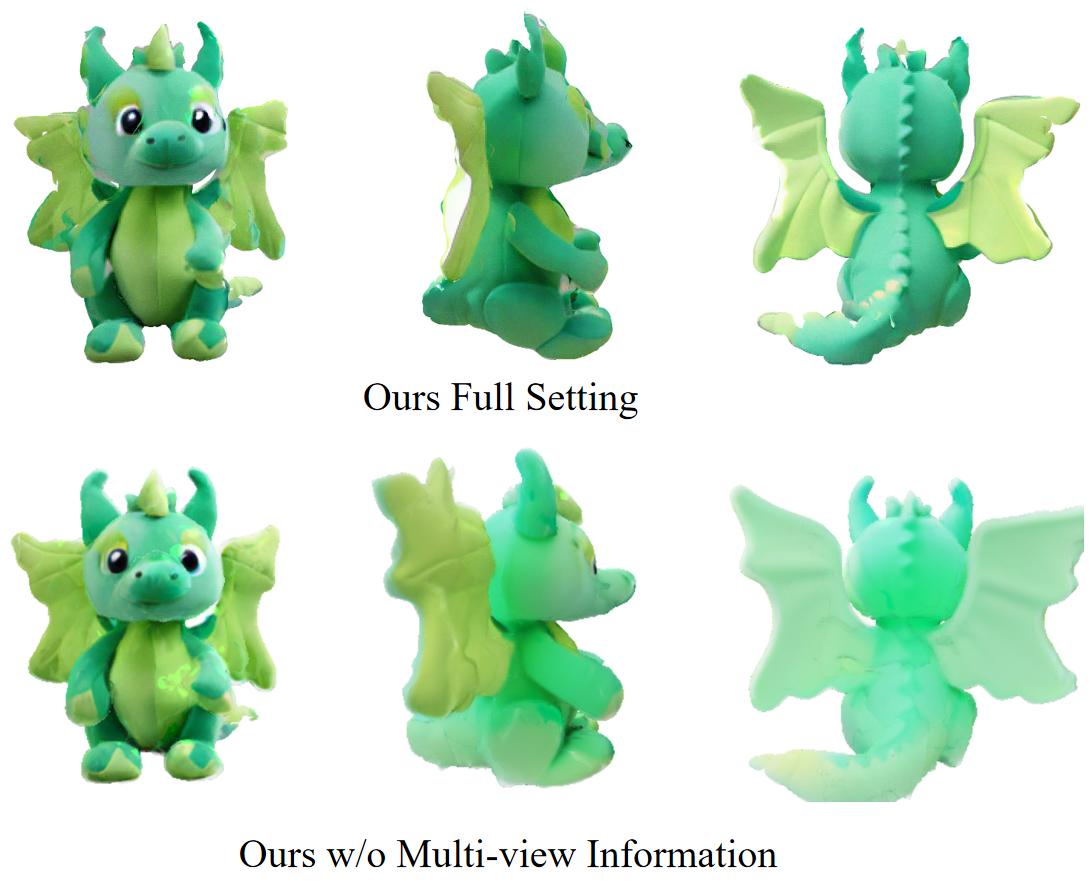}
        \caption{Ablation of Multi-view Information.}
        \label{fig:ablation_mv}
    \end{minipage}
\end{figure}}{\begin{figure}[h]
    \centering
    \begin{minipage}{0.6\textwidth}
        \centering
        \includegraphics[width=\textwidth]{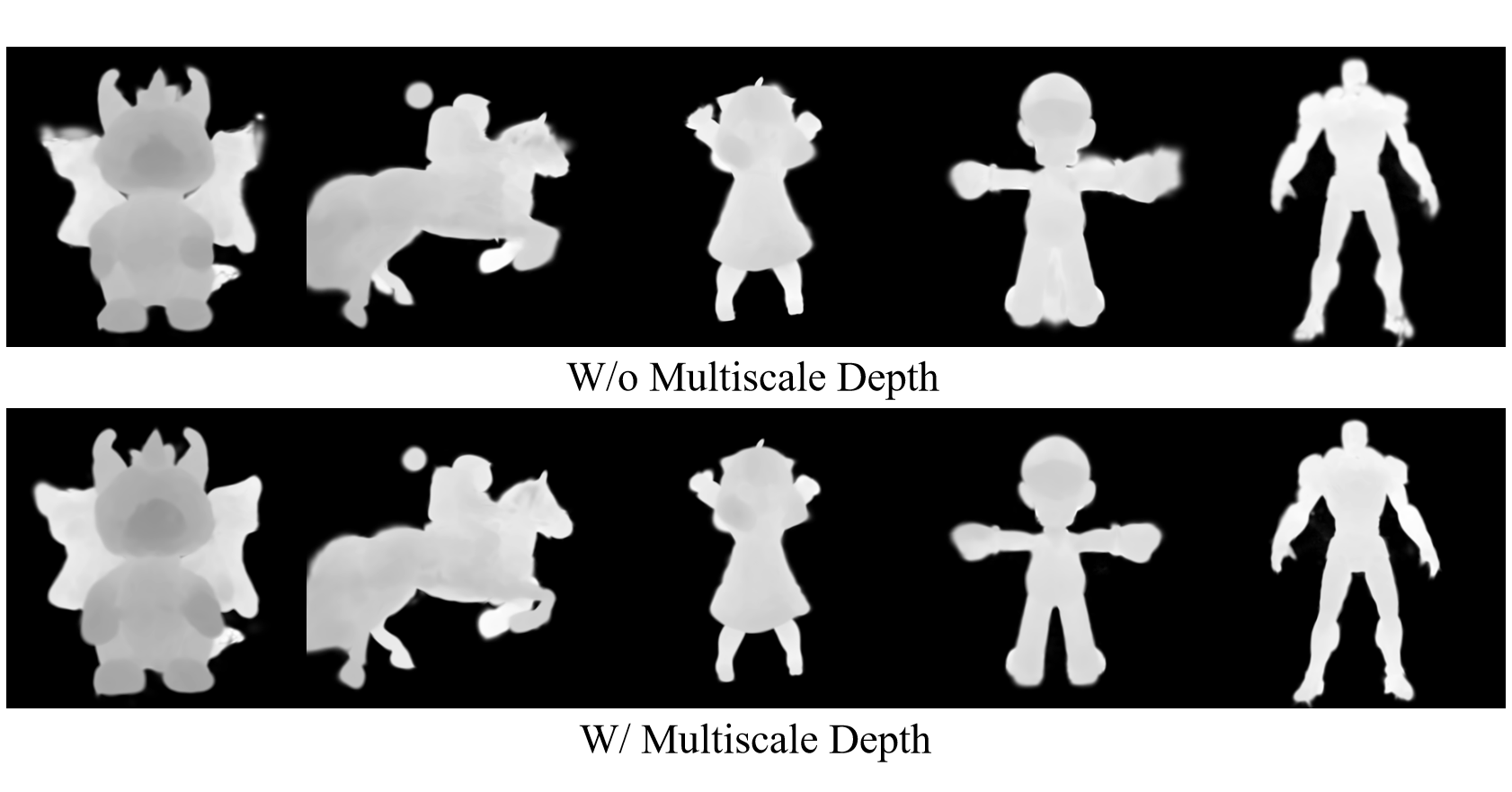}
        \caption{Ablation of Multiscale Depth.}
        \label{fig:ablation_depth}
    \end{minipage}%
    \begin{minipage}{0.4\textwidth}
        \centering
        \includegraphics[width=\textwidth]{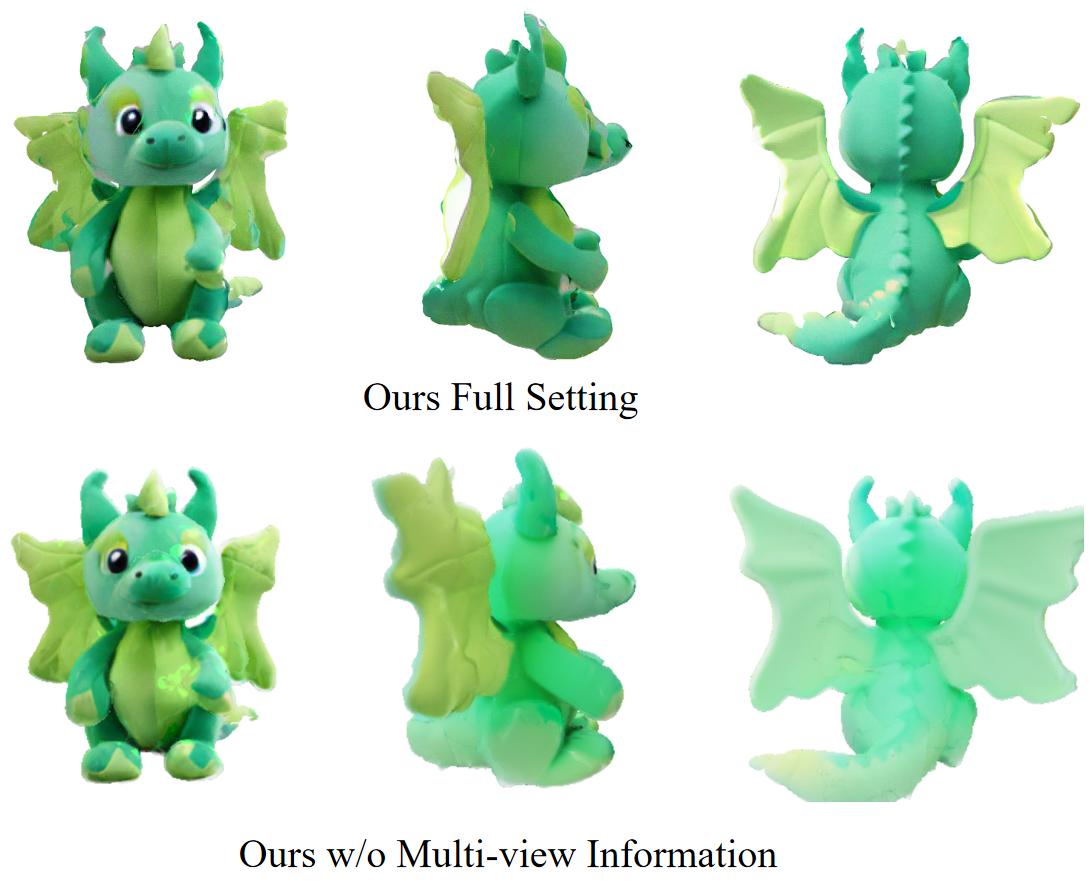}
        \caption{Ablation of Multi-view Information.}
        \label{fig:ablation_mv}
    \end{minipage}
\end{figure}}

\begin{figure}[h]
    \centering
    \includegraphics[width=0.85\textwidth]{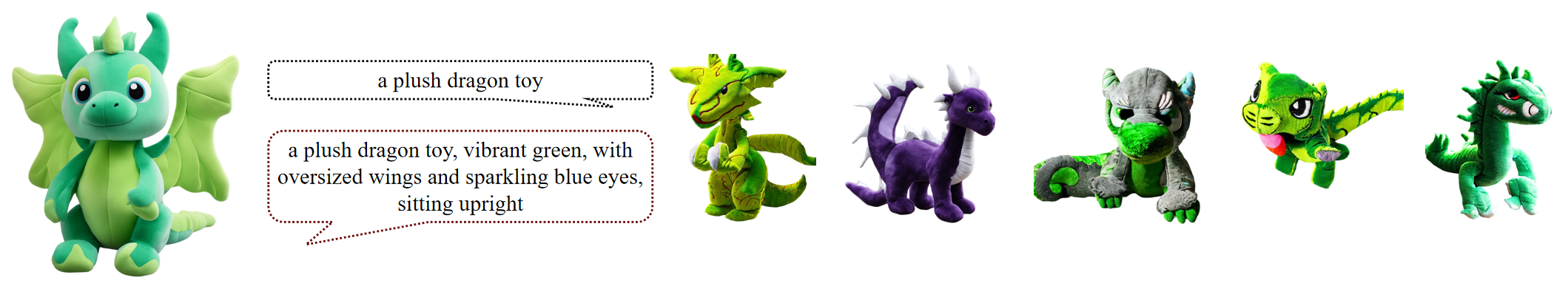}
    \caption{Ablation of Self-Optimization. It is difficult to generate high-quality results with simple text alone. However, by using the multi-modal iterative self-optimization framework, the efficiency of generating high-quality content can be significantly improved.}
    \Description{Ablation of Self-Optimization.}
    \label{fig:ablation_self}
\end{figure}

We further evaluate the importance of the self-optimization process. Since 2D images serve as the foundation for generating high-quality 3D content, they set the upper limit for the visual quality of 3D generation. As shown in the Figure \ref{fig:ablation_self}, the right side displays the results generated using simple user prompts, which often result in low-quality outputs with incomplete content and various issues such as Janus problems. However, with the assistance of GPT-4V, our self-optimization process benefits from timely revisions and quality evaluations, leading to the image on the left. We demonstrate that the self-optimization process, by improving text prompts, significantly enhances image generation quality, thereby impacting the overall quality of the generated 3D content. This capability ultimately helps in selecting more realistic and detailed images from the candidate pool, providing a solid foundation for further 3D content generation.

\section{Conclusion}
In this work, we present an efficient and controllable method for text-to-3D content generation that addresses key challenges in the field. Our approach leverages GPT-4V for self-optimization, enabling the transformation of simple text prompts into richer inputs, thus enhancing generation efficiency and reducing the need for multiple attempts. We further improve controllability by allowing users to specify additional conditions, such as style, edges, scribbles, poses, or combinations of several conditions, for more precise 3D output. By integrating multi-view information during training, we ensure multi-view consistency and improve detail across views, effectively mitigating the Janus problem. Extensive experiments validate that our method achieves robust generalization, offering a practical solution for high-quality, efficient, and controllable 3D content generation.



\appendix

\section{Appendix}

\subsection{More Results}
Since our method can extract mesh from 3D Gaussians, these mesh can be seamlessly applied to downstream tasks, such as rigged animation. The Figure \ref{fig:Ironman_ani} and \ref{fig:spiderman_ani} below show some of our results with rigged animation.
\begin{figure}[h]
    \centering
    \includegraphics[width=0.85\textwidth]{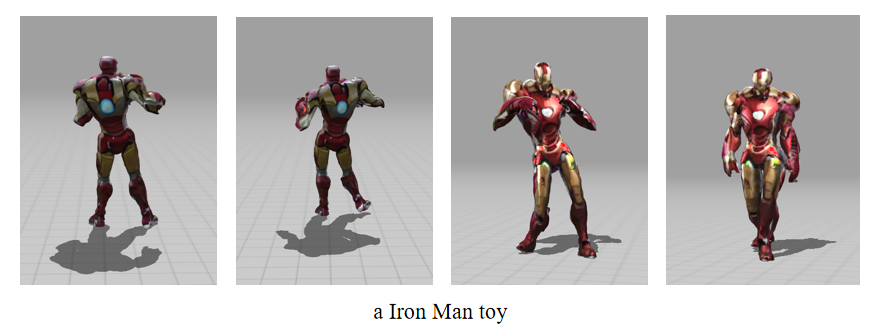}
    \caption{Iron Man toy animation results.}
    \Description{Iron Man toy animation results.}
    \label{fig:Ironman_ani}
\end{figure}
\begin{figure}[h]
    \centering
    \includegraphics[width=0.85\textwidth]{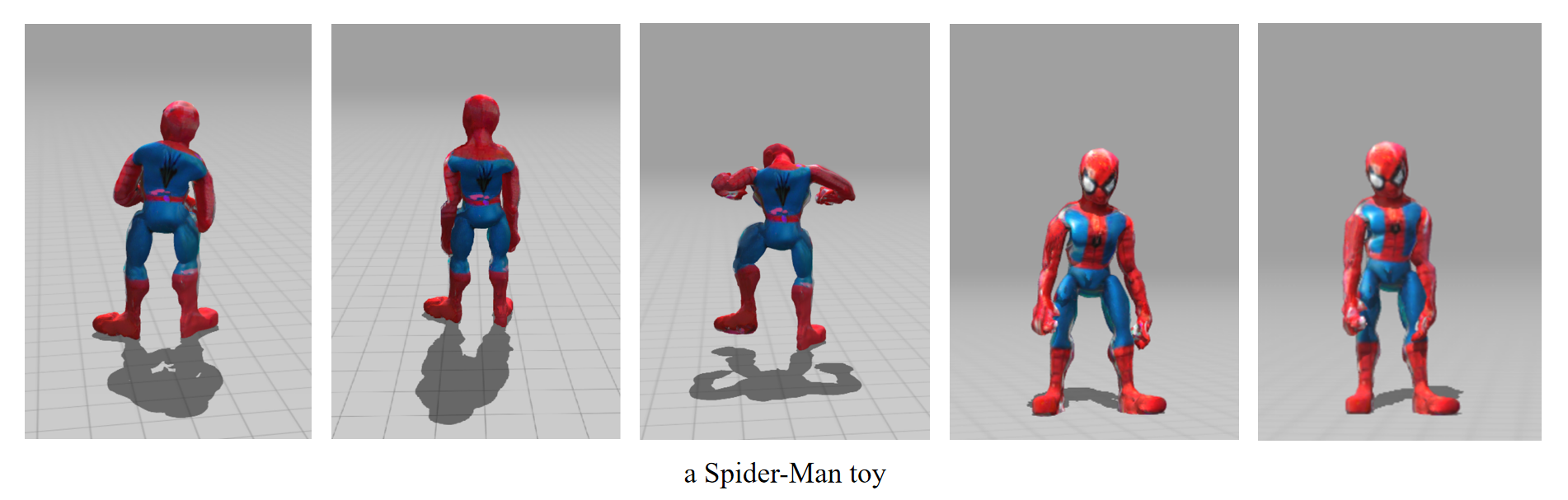}
    \caption{Spider-Man toy animation results.}
    \Description{Spider-Man toy animation results.}
    \label{fig:spiderman_ani}
\end{figure}

\newpage

\bibliographystyle{ACM-Reference-Format}
\bibliography{main}







\end{document}